\documentclass{llncs}

 \usepackage[numbers]{natbib}

\usepackage{graphicx}
\usepackage{amsfonts,amssymb}
\usepackage{paralist}
\usepackage{algorithm}
\usepackage{algorithmic}
\usepackage{epsfig}
\usepackage{array}
\usepackage{makeidx}
\usepackage{chngpage}
\usepackage{color}
\usepackage{url}
\usepackage{makecell}
\usepackage{setspace}
\usepackage{pstricks}
\usepackage{multirow}
\usepackage{hhline}
\usepackage{placeins}
\usepackage{microtype}
\usepackage{longtable}
\usepackage{balance}
\usepackage{tabularx}

\newcolumntype{Z}{>{\centering\let\newline\\\arraybackslash\hspace{0pt}}X}
\newcommand\mathplus{+}

\usepackage{color}

\begin{document}

\mainmatter  

\title{Monkey Optimization System with Active Membranes: A New Meta-heuristic Optimization System}


%
%
\author{Moustafa Zein\inst{1}\and
Aboul Ella Hassanien \inst{1, 4}\and
Ammar Adl\inst{2} \and 
Adam Slowik \inst{3}}
\authorrunning{M. Zein et al.}
%
\institute{Faculty of Computers and Information, Cairo University, Egypt \\
\email{\{moustafazn, aboitcairo\}@gmail.com}\\ \and Faculty of Computers and Information, Beni-Suef University, Egypt 
\email{ammar@fcis.bsu.edu.eg} \and Department of Electronics and Computer Science Koszalin University of Technology, Poland  \email{aslowik@ie.tu.koszalin.pl} \and Scientific Research Group in Egypt (SRGE)}


%
%

\maketitle

\begin{abstract}
Optimization techniques, used to get the optimal solution in search spaces, have not solved
the time-consuming problem. The objective of this study is to tackle the sequential processing problem in Monkey Algorithm and simulating the natural parallel behavior of monkeys. Therefore, a P system with active
membranes is constructed by providing a codification for Monkey Algorithm within the context
of a cell-like P system, defining accordingly the elements of the model - membrane structure,
objects, rules and the behavior of it. The proposed algorithm has modeled the natural behavior
of climb process using separate membranes, rather than the original algorithm. Moreover, it
introduced the membrane migration process to select the best solution and the time stamp was
added as an additional stopping criterion to control the timing of the algorithm. The results
indicate a substantial solution for the time consumption problem, significant representation
of the natural behavior of monkeys, and considerable chance to reach the best solution in the
context of meta-heuristics purpose. In addition, experiments use the commonly used benchmark
functions to test the performance of the algorithm as well as the expected time of the proposed
P Monkey optimization algorithm and the traditional Monkey Algorithm running on population
size $n$. The unit times are calculated based on the complexity of algorithms, where P Monkey
takes a time unit to fire rule(s) over a population size n; as soon as, Monkey Algorithm takes a time unit to run a step every mathematical equation over a population size $n$.

\end{abstract}

\section{Introduction}
Membrane computing is a computational model to simulate and abstract the functionality and structure of biological living cells. It belongs to natural computing science, which investigates the creation of well-designed computational systems based on natural biological systems \cite{252012, 26Zhang_2014}. Illustrations of natural computing contain; neural computation inspired by simulating the functionality of brain, evolutionary computation formulated by the Darwinian evolution of species, swarm intelligence inspired by the behavior of groups of organisms, and life systems modeled by the functionalities of natural life such as membrane computing \cite{27Jiang_2016,elkhani2017membrane,Wenbo2017tissue}. Membrane computing is a theoretical computational model, designed by Gheorghe P\~aun in 1998, providing distributed and parallel computation devices able to solve NP-hard problems in linear or polynomial time \cite{28P_un_2006,Linqiang2018Computational,Bosheng2018Tissue}. Over the past ten years, there were many attractive studies in building connectivity solutions between the potentiality and features of membrane computing and real-life applications, such as membrane algorithms for solving optimization problems \cite{29zhang2014novel,3038Peng_2014,32Peng_2015}. Furthermore, a wide variety of membrane computing applications have been proposed in solving hard computational problems in an efficient way \cite{34Georgiou,35P_rez_Jim_nez,41Peng_2015,garcia2017modeling}. By comparing the results of previous and the existing evolutionary algorithms in membrane computing studies, membrane computing compromises as a more competitive methodology, because of these three advantages: better convergence, stronger robustness and a better balance between exploration and exploitation \cite{3038Peng_2014,37Zhang_2013,39Peng_2015}.

P systems with active membranes and electrical charges were introduced in \cite{7P_un}. This is the first variant in Membrane Computing allowing to solve computationally hard problems in polynomial time. It consists of a hierarchical structure composed of several inner polarized membranes; surrounded by a skin membrane; regions encircled by the membranes containing objects; and evolution rules \cite{940Song_2014,zhang2017data}.
Bio-Inspired swarms and biological systems have been widely used as a computational and optimization solutions for many real applications \cite{42Rathore_2016,43Zhang_2016,gheorghe2017kernel}.  Their models work sequentially, while in reality, swarms' members work together in a governed parallel system. The sequential methodologies used to implement swarms lead to time consumption problem and weaken the power of natural parallel behavior of swarms. One of those swarms models is Monkey Algorithm (MA) \cite{1zhao2008monkey}. It represents the movement of monkeys over mountains to reach the highest mountaintop. Monkeys work in parallel to get into their purpose, but the algorithm does it sequentially. 

This paper aims to enhance the computational efficiency of optimization algorithms by introducing a new algorithm based on a swarm technique called P Monkey System with Active Membranes (PMSAM). It is a new methodology to support the natural parallel behavior of monkeys' movements. It uses the computational power of membrane computing to simulate their movements over mountains, based on a predefined mathematical model. The objective of this paper is to propose a significant solution to time-consuming problem and simulating the natural parallel behavior of monkeys.

The rest of the paper is organized as follows: Section ~\ref{Related Work} introduces P system with active membranes, MA, and its applications. In section ~\ref{Material and methods}, the setup of PMSAM is presented. While section ~\ref{Numerical Experiments and Discussion} reports the outcome of the numerical and empirical experiments. Finally, the main conclusions of the study are summarized and directions for future work are outlined.

\section{Related Work}
\label{Related Work}
In this section, we provide the definitions of P system with active membranes and MA in addition to a discussion about some previous studies that are related to both of them.

\subsection{P system with active membranes}
\label{P system with active membranes}
P system is a research area in computer science targeting the abstraction of the structure and functionality of living cells, in addition to the cells systematized way in tissues or higher cells order structure. Membrane computing has three basic types; cell-like, tissue-like, and neural-like P systems. Cell-like P systems are working on the cell level. P system with active membranes is a class of P systems, which, together with the basic
transition systems and the symport/antiport systems is one of the three
central types of cell-like P systems \cite{7P_un}.

We recall the definition of P system with active membranes that will be used in this paper see \cite{7P_un,8Song_2015,940Song_2014} for more details.

\begin{definition}
	
	A P system with active membranes of degree $m \geq 1$ is a tuple:
	
	$$\Pi = (O, H, \mu, W_1 , ..., W_m, R,0)$$
	Where
	\begin{enumerate}
		\item $O$ is the alphabet of objects;
		\item $H$ is a finite set of labels for membranes;
		\item $\mu$ is a membrane structure consisting of $m$ membranes, injectively labeled with elements of $H$. Each membrane is supposed to have an "electrical polarization", one of the three	possible: positive (+), negative (-), or neutral (0) and is presented as  $S = \{+, -, 0\}$;
		\item $W_1 , ..., W_m$ represents a string of $O$ describing multisets of objects in a region $m$ in $\mu$;
		\item $R$ is a finite set of developmental rules with the following types:
		\begin{enumerate}
			\item $[_h^{s_1} p^v \rightarrow a  ]_h^{s_1}$, \textbf{for} $h\in H,{s_1} \in S, p^v \in O, a \in O^* $
			
			(Object evolution rules; associated with membranes and contingent of the label and the charge of membranes);
			\item $p^{v_1} [_h^{s_1}   ]_h^{s_1}  \rightarrow [_h^{s_2}  p^{v_2} ]_h^{s_2},$  \textbf{for} $h\in H, {s_1}, {s_2}  \in S, p^{v_1}, p^{v_2} \in O $
			
			($in-$communication rules; an object is sent into the membrane, perhaps modified through this process; also the polarization of the membrane can be modified, but not its label);
			\item  $p^{v_1} [_h^{s_1}   ]_h^{s_1}  \rightarrow [_h^{s_2}  ]_h^{s_2}  p^{v_2}$, \textbf{for} $h\in H, {s_1}, {s_2}  \in S, p^{v_1}, p^{v_2} \in O $
			
			($out-$communication rules; an object is sent out of the membrane, perhaps modified through this process; also the polarization of the membrane can be modified, but not its label);
			\item $p^{v_1} [_h^{s_1}   ]_h^{s_1}  \rightarrow  p^{v_2}$, \textbf{for} $h\in H,{s_1}  \in S, p^{v_1}, p^{v_2} \in O $
			
			(Dissolution rules; in reaction with an object, a membrane can be dissolved, while the object specified in the rule can be modified);
			\item $ [_h^{s_1} p^{v_1} ]_h^{s_1}  \rightarrow  [_h^{s_2} p^{v_2} ]_h^{s_2}[_h^{s_3} p^{v_3} ]_h^{s_3}$, \textbf{for} $h\in H,{s_1}, {s_2}, {s_3} \in S, p^{v_1}, p^{v_2}, p^{v_3} \in O $ 
			
			(Division rules for elementary membranes; in reaction with an object,
			the membrane is divided into two membranes with the same label,
			and possibly of different polarizations);
		\end{enumerate}
		\item $0$ is the output region. It represents the output environment.
	\end{enumerate}
\end{definition}

Cell-like P systems are popular paradigm used to solve NP-problems based on their efficiency and computational power such as P system with active membranes \cite{Gheorghe2018dozen,Luis2018From,Alberto2018survey}.  Accordingly, a previous study investigated the computational efficiency of the cell-like P system with Symport/Antiport rules to solve NP-complete problems, QSAT problem, in linear time solution \cite{19Song_2015}. It used some membrane operations (division rules for elementary membranes and communication rule) in the proposed method. The authors further proved that such systems can efficiently solve this complete problem in linear time. A new automatic CNV segmentation method based on an unsupervised and parallel machine earning technique named density cell-like P systems, which injects a clustering algorithm into cell-like P system. Therefore, that study achieved a better accuracy in compassion with previous studies \cite{JIE2018Automatic}. A new study introduced the idea of limited number of membranes to solve satisfiability problem to make P system more realistic model \cite{Yuki2018satisfiability}.

There are two studies introduced the idea of timed P system with active membranes to prove that the correctness of the solution does not depend on the precise timing of involved rules \cite{8Song_2015,940Song_2014}. The authors assumed that firing a rule takes a time unit according to objects interactions. Those studies proved the computational efficiency of those systems with respect to the execution times of inner operations. Also, the types of membranes keep the same initial configurations during the computation. The time unit was calculated from an external clock out of the membrane structure, and it may affect the synchronization between membrane rules and time consumption units.

\subsection{Monkey Algorithm}
Monkey Algorithm is employed to solve global numerical optimization problems with continuous variables \cite{1zhao2008monkey}. The algorithm entails three main processes; Climb, Watch-Jump and Somersault processes, plus the initialization of the algorithm parameters and monkey positions; and the termination process to apply stopping criteria. Climb process finds the local optimal solution in local search space. The Watch-Jump process looks for other points whose objective values exceed those of the current solution. Somersault process makes monkeys find new search spaces rapidly.

	Monkey Algorithm processes are given below \cite{1zhao2008monkey,2Jingran_Wang_2010,3Zheng_2013}:
	\begin{enumerate}
		\item Solution representation:
		\begin{enumerate}
			\item $n$  is the population size of monkeys;
			\item For the monkey $i$, $i \in \{1, \dots ,n\}$ its position is denoted as a vector $p_i=(p_{i 1}, \dots , p_{i d})$ where $d$ is a search space dimension.
		\end{enumerate}

		\item Climb process:
		\begin{enumerate}
			\item  A random vector is generated as
			
			$\Delta p_i=(\Delta p_{i 1}, \dots , \Delta p_{i d})$, where
			\[ \Delta p_{i j} =
			\left\{\begin{array}{lr}
			 a&  with \ probability \  \frac{1}{2}\\
			-a& with \ probability \ \frac{1}{2}
			 \end{array}\right\}  \]
			The parameter $a$ ($a >0$), called the step length of the climb process, can be determined by specific situations.
        	\item For each $i,j, 1 \leq i \leq n, 1 \leq j \leq d$, calculate $f'_{ij} (p_i) = \frac{f(p_i + \Delta p_i) f(p_i - \Delta p_i)}{2 \Delta p_{ij}}$
			
			The vector $ f^{'}_{i} (p_i) = ({f^{'}_{i}}_1(p_i), {f^{'}_{i}}_2(p_i), ..., {f^{'}_{i}}_d(p_i))$ is called the pseudo-gradient of the objective function $f$ at the point $p$;
			\item For each $i, 1 \leq i \leq n$, set $y_i=p_{i j} + a \cdot sign (f'_{i j} (p_i)),$ for $j=1, \dots ,d$, and let $ y = (y_1, y_2, y_3, .., y_d)$;
			\item For each $i, 1 \leq i \leq n$, let $p_i \rightarrow y_i$ if $y_i$ is feasible, otherwise $p_i$ is kept unchanged;
			\item 	Repeat steps a) to d) until there is a little change in the values of objective function in the neighborhood iterations, or the maximum allowable number of iterations (called the climb number, denoted by $N_c$) has been reached.
		\end{enumerate}
		\item Watch-Jump process:
		\begin{enumerate}
			\item For each $j, 1 \leq j \leq d$, randomly generate a real number $y_j \in [p_{i j} - b, p_{i j} + b]$, where $b$ is the eyesight of monkey. Let $y = (y_1, \dots ,y_d)$;
			\item For each $i, 1 \leq i \leq n$, let $p_i \leftarrow (y_1, \dots , y_n)$ if $f(y_1, \dots , y_d) \geq f(p_i)$ and $y=(y_1, \dots , y_d)$ is feasible, otherwise repeat step a) until finding a feasible $y$;
			\item Repeat the climb process by considering $y$ as an initial position.
		\end{enumerate}
		\item Somersault Process:
		\begin{enumerate}
			\item Generate a real random  number $\alpha$ from the interval $[c, g]$ (called the somersault interval);
			\item For each $j, 1 \leq j \leq d$, set $y_j=p_{i j}+\alpha(x_j$ - $p_{i j})$, where $x_j = \frac{1}{n} \sum_{i=1}^n p_{i j}$ and the point
			$x=(x_1, \dots , x_d)$ is called a somersault pivot;
			
			\item For each $i, 1 \leq i \leq n$, let $p_i \leftarrow y$ if $y=(y_1, \dots , y_n)$ is feasible. Otherwise, repeat steps a) and b) until finding a feasible solution $y$.
		\end{enumerate}
		\item Termination Process:
		
	Monkey Algorithm will terminate either after reaching a given number (called the cyclic number, denoted by $N$) of cyclic repetitions of the above steps, or if the optimal value hasn't been changed.
	\end{enumerate}
Monkey Algorithm has many applications and variants in different research fields \cite{Vasundhara2017Behavior}. It's modified to minimize the energy loss, improve the voltage levels, and reduce the carbon dioxide emission. The results show that the proposed algorithm is efficient and achieving good quality solutions \cite{11Duque_2015}.  MA is employed and modified to solve the optimal sensor placement problem. The results show that the modified MA is a better solution than the base one \cite{12zhang2015optimal,13Peng_2016}. The modifications include chaotic initialization, step length, and adaptive watching time. Other many studies introduced modification versions to solve real-life applications such as in \cite{14ZHANG_2016}.
All previous studies focused on enhancing randomization problem with better solutions, but the performance of MA still needs enhancements to overcome the time-consuming problem. No study had proposed a solution for MA enhancement.  Two other studies have introduced an abstract idea about a combination between P system and particle swarm optimization technique \cite{17Singh_2014,18zhou2010particle}. Those studies proved the efficiency and realism of the proposed methodology using benchmark functions. The previous studies neither showed the computational power of P system from the theoretical side nor addressed  P system in swarm processes deeply.

\section{Membrane Monkey Algorithm - Main Definition and Theorems}
\label{Material and methods}
Membrane Monkey Algorithm is responsible for finding the objective value and eliminating the optimal solution of monkey positions. It needs arithmetic operations (addition, subtraction, multiplication, and division), some logical operators (greater than, less than or equal), and ranking algorithm to apply rules of MA processes. We employed previous studies to remove the obstacles of applying the mathematical equations of MA with P system \cite{1zhao2008monkey,2Jingran_Wang_2010,3Zheng_2013}.

\begin{definition}
	
	Membrane Monkey Algorithm of degree $m \geq 1$ is defined as a tuple $$\Pi=(O, H, \mu, W_1, \dots ,W_m, R_g, R_l, 0)$$ 
	
	Where,
	\begin{enumerate}
		\item $O$ is an alphabet of elements called objects. They are corresponding to monkey positions;
		\item $H$ is a set of labels for membranes;
		\item The following membrane structure $\mu = [_g []_{l=1}, []_{l=2},$
		$ []_{l=3}, ..., []_{l=m}]_g$ represents a set of membranes. The skin membrane $[_g]_g$ denotes the global search space, and $ []_{l}$ refers to a local search space (local membrane). These membranes are initially labeled with elements of $H$, where $m$ is the number of regions (local search spaces), $g$ refers to the label of global search space, and $1\leq l \leq m$ is the label of a membrane in the (local) search space. $\mu$ is considered a weighted representation of the natural behavior of monkey movements, through mountains in a large search space to get the highest top mountain. Mountains are local search spaces in large search space named global search space;
		\item The strings $W_i , ..., W_m$ represent multisets of objects placed in the corresponding regions, where $ W = S \epsilon \Pi_{i=1}^{c} p^{v_i}$ for $c\le (n+k)$,  $S$ represents the signal of sending and receiving objects, and $S \in \{-,+,0\}$.  $\epsilon$ represents the message that shows when there is an incoming object to a membrane from another. $n$ indicates the number of monkeys (population size), $k \in \{n, m, d, b, \alpha, f, g, l, p, \Delta Rand, t_{max}, p_c, P_{max} \},$ is a set of algorithm parameters in which defined in Table ~\ref{tab:table1}, and $p^{v_i}$ is an object of the alphabet. The membrane polarization, overall system rules, will not change through firing any rule. It will be the same in both the left and right sides of the rules;
		\item $R_g$  is a finite set of developmental rules in the global search region. It is formulated at Theorem ~\ref{theory1} and ~\ref{theory2};
		
		\item For each $l, 1 \leq l \leq m$, $R_l$ is a region inside a global region. The purpose of those regions is to apply MA processes to get an optimal solution. Each local membrane has a number of rules to perform MA processes (Climb, Watch-Jump and Somersault process). Every local region has the same processes proved at Theorem ~\ref{theory3}, ~\ref{theory4} and ~\ref{theory5};
		
		\item $0$ referred to an environment. The optimal solution and objective value are eliminated to that environment after performing all algorithm processes. This fact is implemented in the model based on some dissolution rules in the termination process.
	\end{enumerate}
\end{definition}
	\begin{table}[!ht]
		\caption{Algorithm Initialization Parameters.}
		\label{tab:table1}
		\centering
		\begin{tabularx}{0.7\linewidth}{| l | Z |}
			\hline
			\textbf{ Parameter   } &  \textbf{ Symbol  }\\
			
			\hline
			Population size&$n$\\
			\hline
			Number of local membranes& $m$\\
			\hline
			Monkey step length&$l$\\
			\hline
			Monkey eyesight&$b$\\
			\hline
			Random real vector&$\Delta Rand$\\
			\hline
			A random real number&$\alpha$\\
			\hline
			Somersault interval values&$[f- g]$\\
			\hline
			Maximum time of algorithm running&$t_{max}$\\
			\hline
			Number of problem dimensions&$d$\\
			\hline
			Number of climb process iterations&$P_c$\\
			\hline
			Number of algorithm iterations&$N_{max}$\\
			\hline
		\end{tabularx}
	\end{table}

	\begin{theorem}
		\label{theory1}
		Initialization process in MA can be performed in one membrane $[_g]_g$.
	\end{theorem}
	\begin{proof}
		Consider a global membrane $[_g]_g$ has a set of rules $R_g$, which is employed to implement the initialization process of MA. We assume that $[_g]_g$ specific responsibilities such as create monkey position objects $p_i$ and algorithm parameters, create local membranes $[_l]_l$ and sending objects to them, and have a full control over the time object. In order to prove that a P system can implement the previous steps in one membrane, we only need to apply $R_g$ in this sequence:
		
		\begin{enumerate}
			\item  Evolution rule:
			
			$[_h^{s_1} p^v \rightarrow a  ]_h^{s_1}$,
			\textbf{for} $h\in H,{s_1} \in S, p^v \in O, a \in O^* $. The rule is responsible for evolving all objects of monkey position and algorithm parameters in the global $[_g ]_g$ and local $[_l ]_l$ regions. Every object $p^v$, $O$ is specified for example real numbers;
			\item $in$-Parameters rule:
			
			$z^{k} [_h^{s_1}   ]_h^{s_1}  \rightarrow [_h^{s_2}  z^{k}   ]_h^{s_2}$,
			\textbf{for} $ h\in H, h =\{h_g, h_l\}, \{s_1, s_2\}  \in S, z^{k}  \in O,$
			$ 1 \leq l \leq m$. Using this rule, the values of Membrane Monkey parameters are passed to their objects in global $[_g ]_g$ and local $[_l ]_l$ regions. This rule is fired in $[_g ]_g$ as a first rule after the evolution rule, and it's fired after the creation rule in $[_l ]_l$;
			
			\item $in$-Timing rule:
			
			$T [_h^{s_1}   ]_h^{s_1}  \rightarrow [_h^{s_2}  T^{t}   ]_h^{s_2}$ , \textbf{for} $h\in H, h =\{h_g, h_l\},\{s_1, s_2\}  \in S, \{T, T^{t}\}  \in O,$
			$1 \leq l \leq m, 0 \leq t \leq  t_{max}$. The rule is fired to inject the consumption timestamp object $T$ into the global and local regions and defined as $T^t$. The time object $T^t$ is introduced to control the time parameter through algorithm processes and changes during these processes;
			
			\item Time-Access rule:
			
			$\epsilon [_h^{s_1} T^{t}   ]_h^{s_1}  \rightarrow [_h^{s_2}  T^{t}_{new}   ]_h^{s_2}$,
			\textbf{for} $h\in H, h =\{h_g, h_l\}, \{s_1, s_2\}  \in S,$
			$ \{ T^{t}, T^{t}_{new}\}  \in O,1 \leq l \leq m, 0 \leq t \leq  t_{max}$.
			
			The rule is used to increment the timestamp for firing every rule in regions. Every firing process sends a message $\epsilon$ to fire the Time-Access to increase timestamp $T^t$ by a time unit as $T^{t}_{new}$. The addition operation is performed on timestamp $T^t$ according to the addition rules in \cite{23Bonchi__2006} with time complexity $(1)$;
			
			\item Division rule:
			
			$ [_{h_g}^{s_1} z^n_1 z^m_2   ]_{h_g}^{s_1}  \rightarrow [_{h_g}^{s_2} \lambda  z^{th}  ]_{h_g}^{s_2}$,
			\textbf{for}
			$h_g\in H, \{s_1, s_2\}  \in S, \{z^n_1, z^m_2 ,  z^{th}\}  \in O$.  The rule is developed to calculate the number of monkeys per local membrane $th$ by dividing the number of monkeys $n$ by number of membranes $m$. The division operation is performed according to the division rules in \cite{23Bonchi__2006}. The division needs two inner membranes inside the local membrane, and the time complexity is based on $z_1^n$. $z^n_1, z^m_2$, and $z^{th}$ are considered constant parameters inside the membrane structure;
			
			\item $in$-Positions rule:
			
			$ P_i^{v_i} [_{h_g}^{s_1}   ]_{h_g}^{s_1}  \rightarrow [_{h_g}^{s_2}   P_i^{v_i}   ]_{h_g}^{s_2},$
			\textbf{for} $h_g\in H, \{s_1, s_2\}  \in S,  P_i^{v_i}   \in O,1 \leq i \leq n$. The rule is developed to modify positions' objects $ p_i^{v_i }$ with positions' values;
			
			\item Membranes-Creation rule:
			
			$\epsilon [_{h_g}^{s_1} z^{m}   ]_{h_g}^{s_1}  \rightarrow [_{h_g}^{s_2} [_{h_{l=1}}^{s_{l+2}}   ]_{h_{l=1}}^{s_{l+2}}, ...,$ $[_{h_{l=m}}^{s_{l+2}}   ]_{h_{l=m}}^{s_{l+2}}    ]_{h_g}^{s_2}$,
			\textbf{fo}r $ \{h_g, h_l\} \in H, \{s_1, s_2, s_{l+2}\}  \in S,$
			$1 \leq l \leq m,  z^{m} \in O$. The rule is developed to generate local membranes (local regions), according to the number of local search spaces $m$ in each iteration $N_{c_{curr}}$. The membrane creation process is done, when an incoming message object  $\epsilon $ reaches $[_g ]_g$;
			
			\item Distribution rule:
			
			$ P_{i}^{v_{i}} [_{h_{l=1}}^{s_{l}}   ]_{h_{l=1}}^{s_{l}}, ...,   [_{h_{l=m}}^{s_{l}}   ]_{h_{l=m}}^{s_{l}} \rightarrow [_{h_{l=1}}^{s_{l+m}} P_{i=1}^{v_{i=1}}, ..., $
			$P_{i=th}^{v_{i=th}}    ]_{h_{l=1}}^{s_{l+m}}, ...,   [_{h_{l=m}}^{s_{l+m}}   P_{i=(m-1)*th}^{v_{i=(m-1)*th}},$
			$..., P_{i=n}^{v_{i=n}}    ]_{h_{l=m}}^{s_{l+m}}$, \textbf{for} $  h_l \in H, \{s_l, s_{l+m}\}  \in S,1 \leq l \leq m,  P_{i}^{v_{i}} \in O$.
			The rule eliminates monkey objects $ P_{i}^{v_{i}}$ from the global region to local regions  based on a threshold object $z^{th}$. Injecting position values  $P_{i}^{v_{i}}$ is considered a message $\epsilon$ to local regions to commence their processes;
			\item Comparison rule:
			
			$T^{t}[_{h_g}^{s_1} T^{t}_{curr}   ]_{h_g}^{s_1}  \rightarrow [_{h_g}^{s_2}  T^{t}_{new}   ]_{h_g}^{s_2}$,
			\textbf{for} $h_g\in H, \{s_1, s_2\}  \in S, \{T^{t}, T^{t}_{curr}, T^{t}_{new}\} $
			$ \in O, 0 \leq t \leq  t_{max}$.
			The rule is used for comparing the current value of time  $T^t_{curr}$  in the global region with incoming time object $T^t$ from a local region. It is fired to update the value of time object $T^{t}_{new}$ in $[_g ]_g$;
			
			\item Time-Updating rule:
			
			$  T^{t}_{new}[_{h_l}^{s_l} T^{t}_{curr}   ]_{h_l}^{s_l}  \rightarrow [_{h_l}^{s_{l+1}}  T^{t}_{new}   ]_{h_l}^{s_{l+1}}$, \textbf{for} $h_l\in H,$
			$\{s_l, s_{l+1}\}  \in S, \{T^{t}, T^{t}_{curr}, T^{t}_{new}\}  \in O,$
			$ 0 \leq t \leq  t_{max}, $
			$1 \leq l \leq m , l \neq g$.
			The rule is developed to update time object $T^t$ in $[_{h_l} ]_{h_l}$ from $[_g ]_g$. It is fired after any change in number of membranes.
		\end{enumerate}
		If there is a P system that has $R_g$ performed with the sequence mentioned and described above, a set of local membranes $[_l]_l$ will be created and PMSAM Time $T$ will be controlled. As a result, the main processes of PMSAM can start.
	\end{proof}

	\begin{theorem}
		\label{theory2}
		The global membrane $[_g]_g$ can get the optimal solution $p_g$ with respect to the maximum number of iterations $N_{max}$ and running time $T_{max}$.
	\end{theorem}
	\begin{proof}
		Let $[_g]_g$ has a responsibility to implement the termination process of PMSAM. To prove that assumption, the stopping criteria (maximum timestamp $T_{max}$, number of algorithm iterations $N_{max}$ or finding a feasible solution for monkey positions and objective value) need to be applied by P system operations. Therefore, a set of rules is constructed based on dissolution, and evolution operations, as follows:
		\begin{enumerate}
			\item $in$-Global rule:
			
			$p_{g} [_{h_g}^{s_1}   ]_{h_g}^{s_1}  \rightarrow [_{h_g}^{s_2} p^{v_{g}}_{g}   ]_{h_g}^{s_2},$ \textbf{for} $h_g\in H,$
			$  \{s_1, s_2\}  \in S,\{ {p_{g}, p^{v_{g}}_{g} } \} \in O$.
			When the global region receives a message $\epsilon$ at the first iteration, the rule is developed to create a global solution object $p^{v_{g}}_{g}$ in the global membrane;
			
			\item Comparison rule:
			
			$p^{v_{opt}}_{opt} [_{h_g}^{s_1} p^{v_{curr}}_{curr}  ]_{h_g}^{s_1}  \rightarrow [_{h_g}^{s_2} p^{v_{g}}_{g}   ]_{h_g}^{s_2}$,
			\textbf{for} $h_g\in H,  \{s_1, s_2\}  \in S,\{ p^{v_{opt}}_{opt}, p^{v_{curr}}_{curr},  p^{v_{g}}_{g}  \}$
			$ \in O$.
			The rule is developed to compare incoming optimal solution object $p^{v_{opt}}_{opt}$ at the current iteration with the current global solution object $p^{v_{curr}}_{curr}$ in global membrane. The result is the better object value $p^{v_{g}}_{g}$ from an incoming object and the current object;
			
			\item Timing rule:
			
			$[_{h_g}^{s_1}P_{i}^{v_{i}} p^{v_{curr}}_{curr} \langle T^{t} T^{t}_{max} \rangle  ]_{h_g}^{s_1}  \rightarrow  P_{i}^{v_{i}} p^{v_{g}}_{g} T^{t}$,  \textbf{for}
			$h_g\in H, s_1  \in S,\{P_{i}^{v_{i}},  p^{v_{g}}_{g}, T^{t} \} \in O,$
			$ 0 \leq t \leq  t_{max}, 1 \leq i \leq n$.
			After global region receives all monkey positions from local membranes, the rule is fired to check if the timestamp $T^t$ exceeds the maximum time $T_{max}^t $, and the optimal solution and objective value objects will be eliminated to the environment. The comparison process between $T^t$, and $T_{max}^t$ is done similar to a sorting algorithm as in \cite{24sburlan2003static} with linear time complexity. Regarding time stopping criterion, it is the first study to put the time factor as a stopping condition;
			\item Dissolution rule:
			
			$[_{h_g}^{s_1}P_{i}^{v_{i}} p^{v_{curr}}_{curr}  ]_{h_g}^{s_1}  \rightarrow  P_{i}^{v_{i}} p^{v_{g}}_{g} $,  \textbf{for} $h_g\in H,$
			$ s_1  \in S,\{P_{i}^{v_{i}},  p^{v_{g}}_{g}\} \in O, 1 \leq i \leq n$. If the current solution $ p^{v_{curr}}_{curr}$ is feasible, the rule eliminates final monkey positions $ P_{i}^{v_{i}}$ and the object value $p^{v_{g}}_{g}$. This occurs when all local membranes finished their work. In PMSAM, the solution feasibility can be checked by creating an object to represent the desired value in the global region;
			
			\item Iteration-Checker rule:
			
			$[_{h_g}^{s_1}P_{i}^{v_{i}} p^{v_{curr}}_{curr} \langle z^{N_{curr}} z^{N_{max}} \rangle  ]_{h_g}^{s_1}  \rightarrow  P_{i}^{v_{i}} p^{v_{g}}_{g} $, \textbf{for} $h_g\in H, s_1  \in S,\{P_{i}^{v_{i}},  p^{v_{g}}_{g}, z^{N_{curr}},
			$
			$z^{N_{max}} \} \in O, 1 \leq i \leq n$.
			The role of this rule is to realize if the current PMSAM iteration  $z^{N_{curr}}$  exceeds the maximum number of iterations $z^{N_{max}}$. In this case, the objective value $p^{v_{g}}_{g} $ monkey position objects $P_{i}^{v_{i}}$ will be eliminated to the environment and stop working in PMSAM. The comparison process between  $z^{N_{curr}}$ and  $z^{N_{max}}$  is done similar to a sorting algorithm as in \cite{24sburlan2003static};
			\item Restarting rule:
			
			$[_{h_g}^{s_1}P_{i}^{v_{i}} p^{v_{curr}}_{curr}  z^{N_{curr}}   ]_{h_g}^{s_1}  \rightarrow [_{h_g}^{s_2} P_{i}^{v_{i}}  z^{N_{new}} ]_{h_g}^{s_2} $, \textbf{for} $h_g\in H, \{s_1, s_2\}  \in S, \{ P_{i}^{v_{i}},   z^{N_{curr}}, $
			$ z^{N_{new}} \} \in O, 1 \leq i \leq n$.
			The rule is used to restart the process from the beginning, in case of the solution is not feasible or the stopping criteria have not been met. The value of current iteration object $N_{curr}$ is incremented by one to be $N_{new} = N_{curr+1}$. It also sends a starter message $\epsilon$ to Membranes-Creation rule to start creating local membranes and passing monkey position objects $P_{i}^{v_{i}}$.
		\end{enumerate}
		
		Formally, from the above rules,  $[_g]_g$ can control PMSAM runtime and the statement of the theorem holds.
	\end{proof}

	\begin{theorem}
		\label{theory3}
		Climb process in PMSAM can be emulated in a homogeneous P system by $m$ membranes where $m$ is the number of local search spaces.
		
	\end{theorem}
		\begin{proof}
		We prove the theorem by finding the local optimal solution in every local membrane $[_l]_l$, which is the purpose of Climb process. Local membranes start to fire climb process rules by generating a random vector according to step length of monkey climbing. After that, they calculate the pseudo-gradient of a selected objective function and finally generate new positions vector to update monkey positions. The set of rules to be fired in $[_l]_l$ are described as follows:
		\begin{enumerate}
			\item Random-Injection rule:
			
			$ \Delta R [_{h_l}^{s_1}   ]_{h_l}^{s_1}  \rightarrow [_{h_l}^{s_2}  \Delta  Rand_{i_u}^{l_{i_u}}   ]_{h_l}^{s_2}$,
			\textbf{for} $h_l\in H, \{s_1, s_2\}  \in S, \{ \Delta R, \Delta  Rand_{i_u}^{l_{i_u}} \}  \in O,$
			 \[
   {l_i}_  = \left\{\begin{array}{lr} l& with \  probability \ge \frac{1}{2}\\
			-l&with \  probability <  \frac{1}{2}\end{array}\right\}\]

		$$	1 \leq i \leq n, 1 \leq l \leq m, 1 \leq u \leq d$$
			The rule passes a random vector $\Delta Rand_{i}^{l_{i}}$ in every local region for all monkeys. The random vector generation is based on the step length of monkey climb $l$ and the dimensionality of the problem.
			\item Addition rule:
			
			$  [_{h_l}^{s_1}  P_i^{v_i} \Delta  Rand_{i}^{l_{i}} ]_{h_l}^{s_1}  \rightarrow [_{h_l}^{s_2}  \Delta P_i^{{v_i}^{'}} ]_{h_l}^{s_2}$,
			\textbf{for} $h_l\in H,$
			$ \{s_1, s_2\}  \in S,
			\{  P_i^{v_i}, \Delta  Rand_{i}^{l_{i}}, \Delta P_i^{{v_i}^{'}} \}$
			$  \in O, 1 \leq i \leq n, 1 \leq l \leq m$.
			The rule is employed to do the first step in calculating the objective function value. When it's fired, an addition process occurs between monkey position value $P_i^{v_i}$ and a random value $\Delta  Rand_{i}^{l_{i}}$ to get a randomized value $ \Delta P_i^{{v_i}^{'}}$ for each monkey position. The addition operation  is performed according to the addition rules in \cite{23Bonchi__2006} with time complexity (1).
			\item Subtraction rule:
			
			$  [_{h_l}^{s_1}  P_i^{v_i} \Delta  Rand_{i}^{l_{i}} ]_{h_l}^{s_1}  \rightarrow [_{h_l}^{s_2}  \Delta P_i^{{v_i}^{''}} ]_{h_l}^{s_2}$, \textbf{for} $h_l\in H, \{s_1, s_2\}  \in S, \{  P_i^{v_i}, \Delta  Rand_{i}^{l_{i}},$
			
			$ \Delta P_i^{{v_i}^{''}} \}  \in O,$
			$ 1 \leq i \leq n,$
			$ 1 \leq l \leq m$
			It is employed to do the second step in calculating an objective function value. When it's fired, a subtraction process occurs between a monkey position value $P_i^{v_i}$  and a random value $\Delta  Rand_{i}^{l_{i}}$ through problem dimensionality $d$ to get a randomized value $ \Delta P_i^{{v_i}^{''}}$ for each monkey position. The subtraction operation is performed according to the subtraction rules in \cite{23Bonchi__2006} with time complexity (1).
			\item Objective rule:
			
			$  [_{h_l}^{s_1} f^{'} \langle \Delta P_i^{{v_i}^{'}} \rangle f^{'} \langle \Delta P_i^{{v_i}^{''}} \rangle ]_{h_l}^{s_1}  \rightarrow [_{h_l}^{s_2} \lambda \Delta P_i^{{v_i}^{'''}} ]_{h_l}^{s_2}$, \textbf{for}
			$h_l\in H, \{s_1, s_2\}  \in S, \{  \Delta P_i^{{v_i}^{'}}, $
			
			$\Delta P_i^{{v_i}^{''}}, \Delta P_i^{{v_i}^{'''}} \}$
			$  \in O, 1 \leq i$
			$ \leq n, 1 \leq l \leq m$. The rule is used to apply an objective function on the result of previous two rules, and it is considered the pseudo-gradient of the objective function. When the rule is fired, a subtraction process occurs between $f^{'} \langle \Delta P_i^{{v_i}^{'}} \rangle$ and  $f^{'} \langle \Delta P_i^{{v_i}^{''}} \rangle$ values  after applying objective function $f^{'}$ rules on them. The purpose of firing this rule is to get a new value $\Delta P_i^{{v_i}^{'''}}$ for each monkey position. To fire this rule, the objective function needs to be applied on $\Delta P_i^{{v_i}^{''}}$ and $ \Delta P_i^{{v_i}^{'}}$, and represented in P system rules.
			
			\item Division rule:
			
			$  [_{h_l}^{s_1}  \Delta P_i^{{v_i}^{'''}} 2 \Delta  Rand_{i_u}^{l_{i_u}} ]_{h_l}^{s_1}  \rightarrow [_{h_l}^{s_2} \lambda  P_i^{{v_i}^{'}} ]_{h_l}^{s_2}$, \textbf{for} $h_l$
			$\in H, \{s_1, s_2\}  \in S, \{  \Delta P_i^{{v_i}^{'''}},$
			
			$ \Delta  Rand_{i_u}^{l_{i_u}},  P_i^{{v_i}^{'}} \}$
			$  \in O, 1 \leq i \leq n,$
			$ 1 \leq l \leq m, 1 \leq u \leq d$. The rule is developed to calculate the pseudo-gradient of the objective function for each monkey position. A new value $P_i^{{v_i}^{'}}$ for each monkey position is calculated by dividing $ \Delta  P_i^{{v_i}^{'''}}$ by $\Delta Rand_{i_u}^{l_{i_u}}$ multiplied by 2. The multiplication is performed according to the multiplication rule in \cite{1zhao2008monkey} with time complexity $O(x log x)$, where $ x$ is the number of bits of the two numbers.
			\item Sign rule:
			
			$  [_{h_l}^{s_1} l {P_i^{{v_i}^{'}}}_{sign}  ]_{h_l}^{s_1}  \rightarrow [_{h_l}^{s_2}  P_i^{{v_i}^{''}} ]_{h_l}^{s_2}$,
			\textbf{for} $h_l\in H,\{s_1, s_2\}  \in S, \{ {P_i^{{v_i}^{'}}}_{sign} ,  P_i^{{v_i}^{''}} \}$
			$  \in O, 1 \leq i \leq n, 1 \leq l \leq m$.
			The rule is used to calculate an updated value $P_i^{{v_i}^{''}}$  of monkey position  based on the pseudo-gradient of an objective function value $P_i^{{v_i}^{'}}$. The  sign function is represented mathematically as:
   \[
   sign(x) = \left\{\begin{array}{lr}	1  &   if \ x > 0\\
			0  &   if \ x =  0\\
			-1 &  if\ x<0\end{array}\right\}\]
  
			The $sign$ function is broken down into a number of P system rules as follows:
			\begin{enumerate}
				\item $  [_{h_l}^{s_1} P_i^{{v_i}^{'}}  ]_{h_l}^{s_1}  \rightarrow [_{h_l}^{s_2} {P_i^{{v_i}^{'}}}_{sign} ]_{h_l}^{s_2}$, \textbf{for} $h_l\in H,$
				$ \{s_1, s_2\}  \in S, \{ P_i^{{v_i}^{'}},  {P_i^{{v_i}^{'}}}_{sign}  \}  \in O,$
				
				$ {P_i^{{v_i}^{'}}}_{sign} =1, P_i^{{v_i}^{'}} >0, 1 \leq i$
				$ \leq n, 1 \leq l \leq m $
				\item $  [_{h_l}^{s_1} P_i^{{v_i}^{'}}  ]_{h_l}^{s_1}  \rightarrow [_{h_l}^{s_2} {P_i^{{v_i}^{'}}}_{sign} ]_{h_l}^{s_2}$, \textbf{for} $h_l\in H,$
				$ \{s_1, s_2\}  \in S, \{ P_i^{{v_i}^{'}},  {P_i^{{v_i}^{'}}}_{sign}  \}  \in O, $
			
				${P_i^{{v_i}^{'}}}_{sign} =0,P_i^{{v_i}^{'}} =0, 1 \leq i$
				$ \leq n, 1 \leq l \leq m $
				\item $  [_{h_l}^{s_1} P_i^{{v_i}^{'}}  ]_{h_l}^{s_1}  \rightarrow [_{h_l}^{s_2} {P_i^{{v_i}^{'}}}_{sign} ]_{h_l}^{s_2}$, \textbf{for} $h_l\in H,$
				$ \{s_1, s_2\}  \in S, \{ P_i^{{v_i}^{'}},  {P_i^{{v_i}^{'}}}_{sign}  \}  \in O, $
			
				${P_i^{{v_i}^{'}}}_{sign} =-1,P_i^{{v_i}^{'}} <0,$
				$ 1 \leq i \leq n,$
				$ 1 \leq l \leq m$. After firing $sign$ rules, the sign-calculation rule is fired.
				
			\end{enumerate}
			
			\item Updating-Position rule:
			
			$  [_{h_l}^{s_1} P_{i_u}^{v_{i_u}} P_i^{{v_i}^{''}}  ]_{h_l}^{s_1}  \rightarrow [_{h_l}^{s_2} y_{i_u} ]_{h_l}^{s_2}$, \textbf{for}  $h_l\in H,$ $\{s_1, s_2\}  \in S, \{ P_{i_u}^{v_{i_u}}, P_i^{{v_i}^{''}}, y_{i_u}   \}  \in O,$
			$ 1 \leq i \leq n, 1 \leq l \leq m, 1 \leq u \leq d $. It is fired to calculate an updated value $ y_{i_u}$ for each monkey position, based on $ P_i^{{v_i}^{''}}$; calculated from the step length of climb process $l$, and the pseudo-gradient of an objective function value $ P_i^{{v_i}^{'}}$.
			
			\item Checker rule:
			
			$  [_{h_l}^{s_1}  y_{i_u} P_{i_u}^{v_{i_u}}  ]_{h_l}^{s_1}  \rightarrow [_{h_l}^{s_2} P_{i_{u-feasible}}^{v_{i_u}} ]_{h_l}^{s_2}$, \textbf{for} $h_l\in H,$
			$\{s_1, s_2\}  \in S, \{   y_{i_u}, P_{i_u}^{v_{i_u}},$
			
			$ P_{i_{u-feasible}}^{v_{i_u}} \}  \in O,$
			$ 1 \leq i \leq n, 1 \leq j$
			$ \leq m, 1 \leq u \leq d$. The rule is developed to compare the updated value $y_{i_u}$ and current value $P_{i_u}^{v_{i_u}}$ of monkey position. It keeps the current value if the updated one is not feasible. The comparison process is done similar to a sorting algorithm as in \cite{24sburlan2003static}, with linear time complexity.
			
			\item Comparison rule:
			
			$  [_{h_l}^{s_1}  y_{i_u} P_{i_u}^{v_{i_u}}  ]_{h_l}^{s_1}  \rightarrow [_{h_l}^{s_2}  y_{i_{u-feasible}} ]_{h_l}^{s_2}$, \textbf{for} $h_l\in H, $
			$\{s_1, s_2\}  \in S, \{   y_{i_u}, P_{i_u}^{v_{i_u}},  y_{i_{u-feasible}} \} $
			
			$ \in O,$
			$ 1 \leq i \leq n, 1 \leq l \leq m,$
			$ 1 \leq u \leq d $.
			The rule is developed to compare updated position value  $y_{i_u}$ and the current value  $P_{i_u}^{v_{i_u}}$ to update current value of monkey positions with the updated value and get a better value $ y_{i_{u-feasible}}$  in process iterations.
			
			\item Repeat steps from the rules in climb process until reaching the maximum number of repeating times of climb process, or get the best local optimal solution.
			
			\item Optimal-Solution rule:
			
			$  [_{h_l}^{s_1} P_{i_{u-feasible}}^{v_{i_u}}  ]_{h_l}^{s_1}  \rightarrow [_{h_l}^{s_2}  P_i^{v_i}  P_{l_{opt}}^{v_{l_{opt}}} ]_{h_l}^{s_2}$, \textbf{for}
			$h_l\in H, $
			$ \{s_1, s_2\}  \in S, \{P_{i_{u-feasible}}^{v_{i_u}},  P_i^{v_i},  $
			
			$P_{l_{opt}}^{v_{l_{opt}}} \}  \in O,$
			$ 1 \leq i \leq n, 1 \leq j$
			$ \leq m, 1 \leq u \leq d$. The rule is developed to get the objective value from the optimal solution for monkey positions. Every local membrane starts to retrieve the local optimal solution $P_i^{v_i}$  and the objective value $P_{l_{opt}}^{v_{l_{opt}}}$ from feasible solution $P_{i_{u-feasible}}^{v_{i_u}}$ by this rule. The objective value $P_{l_{opt}}^{v_{l_{opt}}}$ needs to be ranked as a best solution from monkey positions. The ranking process is similar to the ranking sorting algorithm in \cite{24sburlan2003static} with linear time complexity. There is a little change in ranking sorting algorithm rules; the algorithm eliminates the first element as the best solution and stops working.
			\item Time-Elimination rule:
			
			$ [_{h_l}^{s_1} T^{t}   ]_{h_l}^{s_1}  \rightarrow [_{h_l}^{s_{2}}    ]_{h_l}^{s_{2}}  T^{t}_{out}$,  \textbf{for} $h_l\in H, \{s_1, s_2\}  \in S,$
			$\{T^{t}, T^{t}_{out}\} \in O, 0 \leq t \leq  t_{max}, 1 \leq l \leq m$. The rule is employed to eliminate the time value $T^{t}$ from each local membrane to the global membrane  as $T^{t}_{out}$.
		\end{enumerate}
		The local optimal solution is obtained by the application of previous rules in $[_l]_l$, so the theorem holds.
	\end{proof}
	\begin{theorem}
		\label{theory4}
		The Watch-Jump process can be developed in a homogeneous P system, which gives the optimal solution $y_i$ by migrating $m$ local membranes in one membrane.
	\end{theorem}
	\begin{proof}
		Let us recall the functionality of Watch-Jump process, which allow monkeys to decide whether there are other surrounding points higher than the current one, search for new values exceeding the current solutions and get going with a regeneration rule to create new search domains. To prove the working of above steps in P system, the dissolution operation is applied over $m$ membranes, whereas membranes are merged with each other as follows:
	
		\begin{enumerate}
			\item	Migration rule:
			
			$[_{h_{l=1}}^{s_1}  P_i^{v_i}  P_{l_{opt}}^{v_{l_{opt}}} ]_{h_{l=1}}^{s_1}   [_{h_{l=2}}^{s_2}  P_i^{v_i}  P_{l_{opt}}^{v_{l_{opt}}} ]_{h_{l=2}}^{s_2}$
			$\rightarrow [_{h_{l=q}}^{s_3}  P_i^{v_i}  P_{l_{opt}}^{v_{l_{opt}}} ]_{h_{l=q}}^{s_3}$, \textbf{for} $\{h_l, h_q\}\in H,$
			$\{s_1, s_2, s_3\}  \in S, \{  P_i^{v_i},  P_{l_{opt}}^{v_{l_{opt}}} \}  \in O, $
			$1 \leq i \leq n, 1 \leq l \leq m, 1 \leq q \leq j/2$. The rule is developed to add monkey positions objects $P_i^{v_i}$ from a membrane to another membrane that has a higher optimal solution  $ P_{l_{opt}}^{v_{l_{opt}}}$, where $j = m$ at the first iteration. Migration rule gives a monkey a chance to search for a better position in new search spaces and is fired after finishing climb process. The migration step makes monkeys search for new optimal value and this step avoids the occurring of local minimum.
			
			\item Updating-Position rule:
			
			$  [_{h_q}^{s_1}   P_{i_u}^{v_{i_u}} \langle\pm \rangle z^b  P_{i}^{v_{i}}]_{h_q}^{s_1}  \rightarrow [_{h_q}^{s_2} P_{i}^{v_{i}} y_{i_u} ]_{h_q}^{s_2}$, \textbf{for} $h_q\in H,$
			$\{s_1, s_2\}  \in S, \{   P_{i_u}^{v_{i_u}},   P_{i}^{v_{i}}, y_{i} \}  \in O, 1 \leq i \leq n,$
			$ 1 \leq q\leq j/2, $
			$1 \leq u \leq d, l = q_{max} $. The rule is used to calculate an updated value of monkey position based on a random number, and the eyesight parameter,
			generating a real random number $P_{i_u}^{v_{i_u}} \langle\pm  \rangle  z^b$ based on eyesight $z^b$ of monkeys. Random number generation is done by subtracting or adding the eyesight $z^b$ to the current position to get a new value of monkey position $y_i $.
			
			\item Comparison rule:
			
			$  [_{h_q}^{s_1} f^{'} \langle y_{i_u} \rangle  f^{'} \langle P_{i}^{v_{i}} \rangle]_{h_q}^{s_1}  \rightarrow [_{h_q}^{s_2} y_{i_u} ]_{h_q}^{s_2}$, \textbf{for} $h_q\in H,$
			$\{s_1, s_2\}  \in S, \{    f^{'} \langle y_{i_u} \rangle,  f^{'} \langle P_{i}^{v_{i}} \rangle, y_{i_u} \} $
			
			$ \in O,$
			$ 1 \leq i \leq n, 1 \leq q$
			$\leq j/2, 1 \leq u \leq d, l = q_{max}$. The rule is fired to compare between the current value $P_{i}^{v_{i}}$ and updated value $y_i $ after applying the objective function on both. If an updated value is larger than the current value, the current value is updated.
			
			\item Repeat firing Watch-Jump process rules until q=1.
			
			\item Call and fire climb process rules. The last step in Watch-Jump is to call and fire climb process as in Theorem ~\ref{theory3} until an optimal solution is satisfactory.
		\end{enumerate}
		After applying Watch-Jump process rules, the theorem holds and then moving to Somersault process.
	\end{proof}	
	\begin{theorem}
		\label{theory5}
		A feasible solution $y_i$ is obtained by a set of organized rules in P system; implementing Somersault process.
	\end{theorem}
		\begin{proof}		
		Consider a local membrane $[_l]_l$ has the responsibility of Somersault process; employed to get a new optimal solution and update monkey positions according to somersault pivot. Let us prove this assumption using a membrane that has the following rules:
		\begin{enumerate}
			\item Division rule:
			
			$  [_{h_q}^{s_1}  \langle 1 \langle z^n 1 \rangle \rangle ]_{h_q}^{s_1}  \rightarrow [_{h_q}^{s_2} \lambda z^{n^{'}} ]_{h_q}^{s_2}$,
			\textbf{for} $h_q\in H,\{s_1, s_2\}  \in S, \{ z^n, z^{n^{'}} \}  \in O, q = 1$. The rule is developed to get the value $z^{n^{'}}$ of dividing one by the number of monkeys $z^{n}$. The output of this rule is used to get the somersault pivot.
			
			\item Summation rule:
			
			$  [_{h_q}^{s_1}  \langle P_{i_{u}}^{v_{i_u}}  \rangle_1^n ]_{h_q}^{s_1}  \rightarrow [_{h_q}^{s_2} {P_{{u}}^{v_{u}}}^{'} ]_{h_q}^{s_2}$, \textbf{for} $h_q\in H,\{s_1, s_2\}  \in S, \{  P_{i_{u}}^{v_{i_u}}, {P_{{u}}^{v_{u}}}^{'}\}  \in O,$
			$ q = 1, 1 \leq i \leq n, 1 \leq u \leq d$. The rule is employed to get the total value ${P_{{u}}^{v_{u}}}^{'}$ of summing all monkey position values  $ P_{i_{u}}^{v_{i_u}}$.
			
			\item Pivot rule:
			
			$  [_{h_q}^{s_1}  z^{n^{'}}  {P_{{u}}^{v_{u}}}^{'}  ]_{h_q}^{s_1}  \rightarrow [_{h_q}^{s_2} P_{{u}}^{pv_{u}} ]_{h_q}^{s_2}$, \textbf{for}
			$h_q\in H, \{s_1, s_2\}  \in S, \{ z^{n^{'}},  {P_{{u}}^{v_{u}}}^{'}, P_{{u}}^{pv_{u}}\}  \in O, $
			$q = 1, 1 \leq i \leq n, 1 \leq u \leq d$. The rule is developed to calculate the somersault pivot. When the local membrane receives a message $\epsilon$ about the existence of two objects $z^{n^{'}}$ and  $ {P_{{u}}^{v_{u}}}^{'}$, the rule is fired to calculate somersault pivot $P_{{u}}^{pv_{u}}$ by multiplying $z^{n^{'}}$ and  $ {P_{{u}}^{v_{u}}}^{'}$.
			
			\item Pivot-Process rule:
			
			$  [_{h_q}^{s_1} P_{{u}}^{pv_{u}}  P_{i_{u}}^{v_{i_u}} ]_{h_q}^{s_1}  \rightarrow [_{h_q}^{s_2} \lambda {P_{{u}}^{pv_{u}}}^{'}  ]_{h_q}^{s_2}$, \textbf{for} $h_q\in H,$
			$\{s_1, s_2\}  \in S, \{ P_{{u}}^{pv_{u}},   P_{i_{u}}^{v_{i_u}}, {P_{{u}}^{pv_{u}}}^{'}\}  \in O,$
			$q = 1 , 1 \leq i \leq n, 1 \leq u$
			$\leq d$. The rule is developed to subtract the current position of monkey $P_{i_{u}}^{v_{i_u}}$ from somersault pivot $P_{{u}}^{pv_{u}}$, to get a value ${P_{{u}}^{pv_{u}}}^{'}$ used in calculating the new position.
			
			\item Alfa rule:
			
			$  [_{h_q}^{s_1} {P_{{u}}^{pv_{u}}}^{'} \alpha ]_{h_q}^{s_1}  \rightarrow [_{h_q}^{s_2}{P_{{u}}^{pv_{u}}}^{''} ]_{h_q}^{s_2}$, \textbf{for} $h_q\in H, \{s_1, s_2\}  \in S, \{ {P_{{u}}^{pv_{u}}}^{'}, {P_{{u}}^{pv_{u}}}^{''}\}  \in O, q = 1$. The rule is employed to multiply a new value ${P_{{u}}^{pv_{u}}}^{'}$ obtained from the previous rule with a real random number $\alpha$ to get a new object ${P_{{u}}^{pv_{u}}}^{''}$.
			
			\item New-Positions rule:
			
			$  [_{h_q}^{s_1}  P_{i_{u}}^{v_{i_u}} {P_{u}^{pv_u}}^{''}]_{h_q}^{s_1}  \rightarrow [_{h_q}^{s_2} y_{i_u}  ]_{h_q}^{s_2}$, \textbf{for} $h_q\in H, \{s_1, s_2\}  \in S, \{  P_{i_{u}}^{v_{i_u}} , { P_{{u}}^{pv_{u}}}^{''}, y_{i_u}\}$
			$  \in O, q = 1, 1 \leq i \leq n, 1 \leq u \leq d$. The rule is developed to calculate an updated value of monkey position $y_{i_u}$, by the somersault pivot $P_{i_{u}}^{v_{i_u}}$ and the real random number ${P_{{u}}^{pv_{u}}}^{''}$.
			
			\item Repetition-Checker rule:
			
			$  [_{h_q}^{s_1} y_{i_u}  P_{i_{u}}^{v_{i_u}} ]_{h_q}^{s_1}  \rightarrow [_{h_q}^{s_2} {P_{i_{u}}^{v_{i_u}}}^{'} ]_{h_q}^{s_2}$, \textbf{for} $h_q\in H,$
			$\{s_1, s_2\}  \in S, \{ y_{i_u},  P_{i_{u}}^{v_{i_u}}, {P_{i_{u}}^{v_{i_u}}}^{'}\}  \in O, q = 1,$
			$ 1 \leq i \leq n, 1 \leq u \leq d$.
			The rule is used to check if the optimal solution is not feasible $y_{i_u}$ to repeat somersault process.
			
			\item Call the previous rules of somersault process in case of the optimal solution $y_{i_u}$ is not feasible.
			\item Feasibility-Checking rule:
			
			$  [_{h_q}^{s_1} y_{i_u}  P_{i_{u}}^{v_{i_u}} ]_{h_q}^{s_1}  \rightarrow [_{h_q}^{s_2} \lambda {y_{i_u}}^{'} ]_{h_q}^{s_2}$, \textbf{for} $h_q\in H,$
			$\{s_1, s_2\}  \in S, \{ y_{i_u},  P_{i_{u}}^{v_{i_u}}, {y_{i_u}}^{'}\}  \in O, q = 1,$
			$ 1 \leq i \leq n, 1 \leq u \leq d$.
			The rule is employed to make sure that the new optimal solution $y_{i_u}$ for monkey positions is feasible.
			
			\item Solution-Elimination rule:
			
			$  [_{h_q}^{s_1} {y_{i_u}}^{'} ]_{h_q}^{s_1}  \rightarrow P_{i_{u}}^{v_{i_u}}  P_{q_{u-curr}}^{v_{q_{u-curr}}}$, \textbf{for} $h_q\in H,$
			$\{s_1, s_2\}  \in S, \{  {y_{i_u}}^,P_{i_{u}}^{v_{i_u}},  P_{q_{u-curr}}^{v_{q_{u-curr}}} \}  \in O,$
			$ q = 1 , 1 \leq i \leq n, 1 \leq u \leq d$.
			After performing somersault process, the rule is fired to eliminate the feasible solution ${y_{i_u}}^{'}$  to the global membranes. After the elimination process, the optimal solution (monkey positions $ P_{i_{u}}^{v_{i_u}} $  , and the objective value $P_{q_{u-curr}}^{v_{q_{u-curr}}}$ will be injected into the global membrane.
			\item Time-Elimination rule:
			
			$ [_{h_q}^{s_1} T^{t}   ]_{h_q}^{s_1}  \rightarrow [_{h_q}^{s_{2}}    ]_{h_q}^{s_{2}}  T^{t}_{out}$, \textbf{for} $h_q\in H, \{s_1, s_2\}$
			$ \in S, \{T^{t}, T^{t}_{out}\} \in O,q = 1 , 0 \leq t \leq  t_{max}$.
			The rule is fired to eliminate time value $T^t$ as $T_{out}^t$ from each local membrane to the global membrane, but the local membrane will be dissolved.
		\end{enumerate}
		
		As a result, the global membrane can get a feasible solution for every iteration and determine the optimal solution, so the theorem holds.
	\end{proof}	
	
The progress of PMSAM and the sequence of firing algorithm rules is shown in Algorithm~~\ref{algorithm1}. The algorithm starts to get the input parameters and evolve the need objects. After that, the Climb, Watch-Jump, and Somersault processes are performed based on PMSAM theorems. At every iteration, the algorithm checks the stopping criteria.

\begin{algorithm}
	\footnotesize
	\caption{Pseudo-code algorithm of PMSAM}
	\label{algorithm1}
\begin{algorithmic}[1]
	\REQUIRE  $\mu = [_g]_g$, $\Delta R$, $p$ and  $k$   \COMMENT{\textit{P system input configuration.}}
	\ENSURE   $p_g$ and  $T$ \\
	\STATE $N \gets 1$, $m \gets 1$ and $T \gets 0$
	\STATE \textit{Step 1:} \textbf{evolve} objects \textbf{by} \textit{Evolution rule}
	\STATE \textit{Step 2:} \textbf{fire} \textit{$in$-Parameters, $in$-Positions and $in$-Timing rules}
	\STATE \textit{Step 3:} $T \gets 1$ \textbf{by} \textit{Time-Access rule}
	\WHILE{$N \le N_{max}$ $\AND$ $T \le T_{max}$}
	\STATE \textbf{create} $m$ membranes \textbf{by} \textit{Division, Membranes-Creation and Time-Access rules}
	\STATE \textbf{send} $p$ and $k$ to $m$ membranes \textbf{by} \textit{$in$-Parameters, Distribution, $in$-Timing and Time-Access rules}
	\WHILE{ $i \le p_{c}$  \COMMENT{\textit{check maximum number of climb iteration inside local membranes in parallel}}}
	\STATE \textbf{fire} \textit{Evolution, Random-Injection and Time-Access rules}
	\STATE	\textbf{apply} \textit{Objective function and Time-Access rules}
	\STATE \textbf{update} $p$ \textbf{by} \textit{Updating-position and Time-Access rules}
	\STATE \textbf{update} $p_{feasible}$ \textbf{by} \textit{Comparison and Time-Access rules}
	\STATE \textbf{set} $i = i + l$ \textbf{by} \textit{checker and Time-Access rules}
	\ENDWHILE
	\STATE \textbf{set} $p_{g}$ \textbf{with} $p_{feasible}$ \textbf{by} \textit{optimal-Solution and Time-Access rules}
	\STATE \textbf{fire} \textit{Time-Access and Time-Elimination rules}
	\STATE \textbf{update} $T_{new}$ \textbf{with} $T$ \textbf{by} \textit{Timing and Time-Access rules}
	\WHILE{$l\ge 2$ \COMMENT{\textit{continue firing rules until number of local membranes = 1}}}
	\STATE \textbf{dissolve} in parallel $[_l]_l[_l]_l$ into $[_q]_q$ \textbf{by} \textit{Migration and Time-Access rules}
	\STATE \textbf{inject} and \textbf{fire} Watch-Jump rules into $[_q]_q$   \COMMENT {\textit{Call Theorem ~\ref{theory4}}}
	\ENDWHILE
	\WHILE{$p$ is not feasible \COMMENT{\textit{Feasibility-Checking, Repetition-Checker and Time-Access rules}}}
	\STATE \textbf{inject} and \textbf{fire} Somersault rules into $[_q]_q$  \COMMENT{\textit{Call Theorem ~\ref{theory5}}}
	\ENDWHILE
	\STATE \textbf{dissolve} $p$ and $T$ \textbf{by} \textit{Solution-Elimination, Time-Elimination and Time-Access rules}
	\STATE \textbf{fire} \textit{Global, Time-Updating and Time-Access rules}
	\STATE \textbf{set} $N = N + l$ \textbf{by} \textit{Time-Access, Iteration-Checker, Restarting and Timing rules}
	\ENDWHILE
	\RETURN  $p_{g}$ \textbf{by} \textit{Dissolution rule}
\end{algorithmic}
\end{algorithm}

\section{Experiments and Discussion}
\label{Numerical Experiments and Discussion}
In this section, we imply two different approaches of PMSAM experiments. The first approach is a theoretical  trace to prove that the proposed algorithm is halt and reaches the success case. On the other hand, there are experiments to evaluate the time complexity, convergence and optimal solution.
\subsection{Numerical  Experiment}
A numerical example is introduced to prove the computational power and efficiency of the proposed algorithm and how it halts successfully. The following experiment starts with population initialization (monkeys, and algorithm parameters). Let us suppose that a monkey population has $n = 20$ monkeys and $m = 4$ membranes initialized as $$p^{v_i }  =\{ (v_1,rand_1 ), (v_2, rand_2 ), ..., (v_{20},  rand_{20} ) \}$$  where  $rand$ is a random number. Also, the algorithm parameters were initialized as $$z^k = \{ (n, 20), (m,4), d, b, l, f, g, \alpha, p, t_{max}\}$$
Table ~\ref{tab:table2} shows the workflow of PMSAM among regions (global and local regions), and examines the changes in variable objects (member of membranes $Mem.$, monkey positions $M pos.$ and timestamp $t$) over the previous population. The progress of algorithm steps starts with the initialization process at which, the global region receives messages about incoming objects. It fires the evolution rule to evolve these objects (algorithm parameters and initial monkey positions). Then, the objects convey their values of the algorithm parameters (monkey position values, algorithm parameters and time object in parallel). At that time, the Division rule is fired to compute a threshold value and access time object for the first time to increase it by a one-time unit. Whenever a rule is fired throughout the algorithm, the Time-Access rule is fired to increment the time value by a unit.
Membranes-Creation rule is fired to create four local regions according to the number of local membranes $m$. After that, three communication rules are fired in parallel to send objects (time object, positions, and algorithm parameters) to local membranes according to the threshold.
		
Local membranes creation and objects assignment is the trigger of starting the Climb process in every local membrane in parallel as in Table ~\ref{tab:table2}. All of them start to fire their rules according to climb process rules sequence. The goal of this process is reaching the local optimal solution in every region. In Table ~\ref{tab:table2}, Climb process is based on applying an objective function, that needs to map its mathematical formula to P system rules. For example, the following objective function $x^2$ is mapped to be:
	\begin{equation}
	[_h^{s_1} x x ]_h^{s_1}  \rightarrow [_h^{s_2} x^{'} ]_h^{s_2} ,   h\in H, \{s_1, s_2\}  \in S, \{ x,  x^{'}\}  \in O
	\end{equation}
in Equation 1, the rule needs a time unit to be fired, so the time increasing in this step depends on the breakdown of the objective function. After that, the sign function is calculated with the inner number of rules, and monkey positions are updated. The climb process will continue to run until it reaches the maximum number of processes. At every iteration, a comparison rule is fired to update the current solution with the calculated optimal solution in that iteration.  The last step in climb process is eliminating the time object in the global membrane, to update the timestamp in global and local membranes. Before updating time object in local membranes, Migration rule is fired to merge monkeys in two membranes to be in one membrane and start the Watch-Jump process. After the migration process, there are two local membranes instead of four at the first step in Watch-Jump process as in Table ~\ref{tab:table3}.
    
Watch-Jump process continues to work with updating monkey position, applying the objective function again, comparing monkey positions with updated objects, and all the previous steps are repeated once again to get a number of local membranes equal to one. Watch-Jump process calls climb process again, then, the progress of PMSAM is moved to somersault process. It starts with calculating somersault pivot by firing a number of rules to update monkey positions. 
	\begin{table}[H]
		\begin{adjustwidth}{-.5in}{-.5in}
			\centering
			\renewcommand{\arraystretch}{1.9}
			\caption{The first part of the tracing scenario for a numerical example of Membrane Monkey Algorithm.}
			\label{tab:table2}
			\Large
			\resizebox{\textwidth}{!}{\begin{tabular}{|l|c|c|c|c|c|}
					\hline
					\multirow{2}{*}{ \textbf{Global mem.}} &	 \multirow{2}{*}{  \textbf{Local mem.}} &\multicolumn{3}{c}{\textbf{Variables}} \vline&  \multirow{2}{*}{ \textbf{Rule(s)}}\\
					\cline{3-5}
					&&$Mem.$&$M pos.$&$t$&\\
					\hline
					\multicolumn{6}{|c|}{\textbf{Initialization Process: Environment $(P_{1}^{v_{1}}, ..., P_{20}^{v_{20}}, z^k)$}}\\
					\hline
					$p_i z^k T^t [_g]_g$&&5&\makecell{$P_{i}^{v_{i}} =$ $ P_{1}^{v_{1}}, ...,$ $ P_{20}^{v_{20}}$} &\multirow{2}{*}{0}&Evolution\\
					\cline{1-4} \cline{6-6}
					$[_g P_{i}^{v_{i}} z^k T^t]_g$&&5&$P_{i}^{v_{i}}$&&\makecell{$in-$Parameters\\$in-$Timing, $in-$Positions }\\
					\cline{1-6}
					$[_g P_{i}^{v_{i}} z^k z^{th} T^t]_g$&&5&$P_{i}^{v_{i}}$&1&\makecell{Division, Time-Access}\\
					\cline{1-6}
					$[_g P_{i}^{v_{i}} z^k  T^t [_l]_l]_g$&$[_{l=1}]_{l=1} [_{l=2}]_{l=2} [_{l=3}]_{l=3} [_{l=4}]_{l=4}$&5&$P_{i}^{v_{i}}$&2&\makecell{Membranes-Creation\\Time-Access }\\
					\cline{1-6}
					$[_g  z^k  T^t [_l]_l]_g$&\makecell{$[_{l=1} z^k  T^t P_{1}^{v_{1}}, ...,   P_{5}^{v_{5}}]_{l=1},...,$ \\ $ [_{l=4} z^k  T^t P_{16}^{v_{16}}, ..., P_{20}^{v_{20}}]_{l=4}$}&5&$P_{i}^{v_{i}}$&3&\makecell{$in-$Parameters, $in-$Timing\\Distribution, Time-Access }\\
					\hline
					\multicolumn{6}{|c|}{  \textbf{Climb Process}}\\
					\cline{1-6}
					$[_g  z^k  T^t [_l]_l]_g$&\makecell{$\Delta Rand$$[_{l=1} z^k  T^t P_{i}^{v_{i}}]_{l=1}, ...,$\\ $ [_{l=4} z^k T^t  P_{i}^{v_{i}}]_{l=4}$}&5&$P_{i}^{v_{i}}$&4&\makecell{Evolution\\Time-Access }\\
					\cline{1-6}
					$[_g  z^k  T^t [_l]_l]_g$&\makecell{$[_{l=1}\Delta Rand z^k  T^t P_{i}^{v_{i}}]_{l=1},...,$ \\ $ [_{l=4}\Delta Rand z^k  T^t  P_{i}^{v_{i}}]_{l=4}$}&5&$P_{i}^{v_{i}}$&5&\makecell{Random-Injection\\Time-Access }\\
					\cline{1-6}
					$[_g  z^k  T^t [_l]_l]_g$&\makecell{$[_{l=1} f^{'} \langle \Delta {P_{{i}}^{v_{i}}}^{'} \rangle f^{'} \langle \Delta {P_{{i}}^{v_{i}}}^{''} \rangle z^k  T^t P_{i}^{v_{i}}]_{l=1}, ...,$\\  $[_{l=4} f^{'} \langle \Delta {P_{{i}}^{v_{i}}}^{'} \rangle f^{'} \langle \Delta {P_{{i}}^{v_{i}}}^{''} \rangle z^k  T^t  P_{i}^{v_{i}}]_{l=4}$}&5&$P_{i}^{v_{i}}$&7&\makecell{Square-function\\Time-Access }\\
					\cline{1-6}
					$[_g  z^k  T^t [_l]_l]_g$&\makecell{$[_{l=1} \Delta {P_{{i}}^{v_{i}}}^{'''} z^k  T^t P_{i}^{v_{i}}]_{l=1}, ...,$\\  $[_{l=4} \Delta {P_{{i}}^{v_{i}}}^{'''} z^k  T^t  P_{i}^{v_{i}}]_{l=4}$}&5&$P_{i}^{v_{i}}$&8&\makecell{Objective\\Time-Access }\\
					\cline{1-6}
					$[_g  z^k  T^t [_l]_l]_g$&\makecell{$[_{l=1} {P_{{i}}^{v_{i}}}^{''} z^k  T^t P_{i}^{v_{i}}]_{l=1}, ...,$\\  $[_{l=4} {P_{{i}}^{v_{i}}}^{''} z^k  T^t  P_{i}^{v_{i}}]_{l=4}$}&5&$P_{i}^{v_{i}}$&11&\makecell{Sign\\Time-Access}\\
					\cline{1-6}
					$[_g  z^k  T^t [_l]_l]_g$&\makecell{$[_{l=1} y_{i_1} z^k  T^t P_{i}^{v_{i}}]_{l=1}, ...,$\\  $[_{l=4} y_{i_1} z^k  T^t  P_{i}^{v_{i}}]_{l=4}$}&5&$P_{i}^{v_{i}}$&12& \makecell{Updating-Position\\Time-Access}\\
					\cline{1-6}
					$[_g  z^k  T^t [_l]_l]_g$&\makecell{$[_{l=1} y_{i_1} z^k  T^t ]_{l=1}, ...,$\\  $[_{l=4} y_{i_1} z^k  T^t ]_{l=4}$}&5&$ y_{i_1}$&14& \makecell{Optimal-Solution\\Time-Access}\\
					\cline{1-6}
					$[_g  z^k  T^t [_l]_l]_g$&\multicolumn{3}{c}{Repeat the previous steps in climb process until reaching $z^{P_c}$} \vline&120& Time-Access\\
					\cline{1-6}
					$[_g  z^k T^t_{new} T^t [_l]_l]_g$&\makecell{$[_{l=1} P_{i}^{v_{i}} z^k  T^t ]_{l=1}, ...,$\\  $[_{l=4} P_{i}^{v_{i}} z^k T^t  ]_{l=4}$}&5&$P_{i}^{v_{i}}$&121& \makecell{Time-Elimination\\Time-Access}\\
					\hline
					\multicolumn{6}{|c|}{\textbf{Initialization} \& \textbf{Watch-Jump Processes}}\\
					\hline
					$[_g  z^k  T^t [_l]_l]_g$&\makecell{$[_{l=1} P_{i}^{v_{i}} z^k  T^t ]_{l=1} $\\  $[_{l=2} P_{i}^{v_{i}} z^k T^t  ]_{l=2}$}&3&$P_{i}^{v_{i}}$&\makecell{122\\($T^t = $$T^t_{new}$)}& \makecell{Time-comparison\\Migration, Time-Access}\\
					\hline
					\multicolumn{6}{|c|}{\textbf{Initialization Process}}\\
					\hline
					$[_g  z^k  T^t [_l]_l]_g$&\makecell{$[_{l=1} P_{i}^{v_{i}} z^k  T^t ]_{l=1} $\\  $[_{l=2} P_{i}^{v_{i}} z^k T^t  ]_{l=2}$}&3&$P_{i}^{v_{i}}$&\makecell{123\\($T^t = $$T^t_{new}$)}& \makecell{Time-Updating\\Time-Access}\\
					\hline
			\end{tabular}}
		\end{adjustwidth}
	\end{table}

Two rules are fired, in parallel, to check the feasibility of an optimal solution. In Table ~\ref{tab:table3}, the algorithm finds the optimal solution from the first time, so it moved to fire time and solution elimination rules. Otherwise, the somersault process is repeated from the beginning until finding the optimal solution.

In the termination process, stopping criteria depend on three objects (maximum running time, maximum number of iterations, and feasible solution). Stopping criteria have three rules fired to eliminate the optimal solution to the environment and stop working in PMSAM. In case of the solution is not feasible and the current time object does not exceed the maximum time object, or the current iteration number does not exceed the maximum number of iteration, the global region will start the process from the beginning (create membranes and assign monkey positions). In that example, the algorithm has stopped working because the current solution is feasible with time object value $261$ time unit.

\begin{table}[H]
	\begin{adjustwidth}{-.5in}{-.5in}
		\centering
		\renewcommand{\arraystretch}{1.7}
		\caption{The second part of the tracing scenario for a numerical example of Membrane Monkey Algorithm.}
		\label{tab:table3}
		\Large
		\resizebox{\textwidth}{!}{\begin{tabular}{|c|c|c|c|c|c|}
				\hline
				\multirow{2}{*}{ \textbf{Global mem.}} &	 \multirow{2}{*}{  \textbf{Local mem.}} &\multicolumn{3}{c}{\textbf{Variables}} \vline&  \multirow{2}{*}{ \textbf{Rule(s)}}\\
				\cline{3-5}
				&&$Mem.$&$M pos.$&$t$&\\
				\hline
				\multicolumn{6}{|c|}{\textbf{Watch-Jump Process}}\\
				\hline
				$[_g  z^k  T^t [_l]_l]_g$&\makecell{$[_{l=1}y_{i_1} P_{i}^{v_{i}} z^k  T^t ]_{l=1} $\\  $[_{l=2}y_{i_1} P_{i}^{v_{i}} z^k T^t  ]_{l=2}$}&3&$P_{i}^{v_{i}}$&124& \makecell{Updating-Position\\Time-Access}\\
				\cline{1-6}
				$[_g  z^k  T^t [_l]_l]_g$&\makecell{$[_{l=1} f^{'} \langle y_{i_1} \rangle f^{'} \langle P_{i}^{v_{i}} \rangle z^k  T^t ]_{l=1} $\\  $[_{l=2} f^{'} \langle y_{i_1}\rangle f^{'} \langle P_{i}^{v_{i}} \rangle z^k T^t  ]_{l=2}$}&3&$P_{i}^{v_{i}}$&125& \makecell{Objective\\Time-Access}\\
				\cline{1-6}
				$[_g  z^k  T^t [_l]_l]_g$&$[_{l=1}  P_{i}^{v_{i}} z^k  T^t ]_{l=1}$&2&$P_{i}^{v_{i}}$&127& Migration, Time-Access\\
				\cline{1-6}
				$[_g  z^k  T^t [_l]_l]_g$&$[_{l=1}y_{i_1} P_{i}^{v_{i}} z^k  T^t ]_{l=1} $&2&$P_{i}^{v_{i}}$&128&Updating-Position, Time-Access\\
				\cline{1-6}
				$[_g  z^k  T^t [_l]_l]_g$&$[_{l=1} f^{'} \langle y_{i_1} \rangle f^{'} \langle P_{i}^{v_{i}} \rangle z^k  T^t ]_{l=1}$&2&$P_{i}^{v_{i}}$&129&Objective, Time-Access\\
				\cline{1-6}
				$[_g  z^k  T^t [_l]_l]_g$&$[_{l=1}  P_{i}^{v_{i}} z^k  T^t ]_{l=1} $&2&$P_{i}^{v_{i}} = y_{i_1}$&130& Comparison,  Time-Access\\
				\cline{1-6}
				$[_g  z^k  T^t [_l]_l]_g$&\multicolumn{3}{c}{Repeat climb process} \vline&252&  Time-Access\\
				\hline
				\multicolumn{6}{|c|}{\textbf{Somersault Process}}\\
				\hline
				$[_g  z^k  T^t [_l]_l]_g$& $[_{l=1}  P_{{u}}^{pv}  P_{i}^{v_{i}} z^k  T^t ]_{l=1} $&2&$P_{i}^{v_{i}}$&254& Pivot, Time-Access\\
				\cline{1-6}
				$[_g  z^k  T^t [_l]_l]_g$&$[_{l=1}  {P_{{u}}^{pv}}^{'}  P_{i}^{v_{i}} z^k  T^t ]_{l=1} $&2&$P_{i}^{v_{i}}$&255& Pivot-Process, Time-Access\\
				\cline{1-6}
				$[_g  z^k  T^t [_l]_l]_g$&$[_{l=1}  {P_{{u}}^{pv}}^{''}  P_{i}^{v_{i}} z^k  T^t ]_{l=1} $&2&$P_{i}^{v_{i}}$&256& Alfa, Time-Access\\
				\cline{1-6}
				$[_g  z^k  T^t [_l]_l]_g$&$[_{l=1}  y_{i_1}  P_{i}^{v_{i}} z^k  T^t ]_{l=1} $&2&$P_{i}^{v_{i}}$&257& New-Position, Time-Access\\
				\cline{1-6}
				$[_g  z^k  T^t [_l]_l]_g$&$[_{l=1}  {y_{i_1}}^{'}   z^k  T^t ]_{l=1} $&2&$P_{i}^{v_{i}} = y_{i_1}$&258& \makecell{Repetition-Checker\\Feasibility-Checking, Time-Access}\\
				\cline{1-6}
				\makecell{$T^t_{out} P_{opt}^{v_{opt}}[_g $$  z^k  T^t  P_{i}^{v_{i}} P_{curr}^{v_{q_{curr}}}]_g$}&$  T^t_{out}$&1&$P_{curr}^{v_{q_{curr}}} $&259& \makecell{Solution-Elimination\\Time-Elimination, Time-Access}\\
				\hline
				\multicolumn{6}{|c|}{\textbf{Termination} \& \textbf{Initialization Processes}}\\
				\hline
				\makecell{$[_g  z^k  T^t z^{N_{curr}}$$  z^{N_{max}} P_{g}^{v_{g}}]_g$}&&1&$ P_{g}^{v_{g}}$&\makecell{260\\($T^t = T^t_{out}$)}& \makecell{Comparison\\Iteration-Checker, Time-Access}\\
				\hline
				\multicolumn{6}{|c|}{\textbf{Termination Process: Environment $(P_{i}^{v_{i}} P_{g}^{v_{g}})$}}\\
				\hline
				&&0&&261& \makecell{Timing, Dissolution}\\
				\hline
		\end{tabular}}
	\end{adjustwidth}
\end{table}

\subsection{Empirical experiment}
PMSAM is implemented using Java parallel programming features (ExecutorService, ForkandJoin framework and streams). The implementation is just a simulation for P system over the current CPUs and is not a true implementation because there is no any device can work with P system. The need of performing experiments on the proposed algorithm makes the possibility of presenting it by the current devices and environments is valid and acceptable. PMSAM is benchmarked on benchmark functions reported in many previous works \cite{1zhao2008monkey,digalakis2001benchmarking}, and listed in Table ~\ref{tab:table4}.

\begin{table}[H]
	\begin{adjustwidth}{-.5in}{-.5in}
		\centering
		\setlength{\extrarowheight}{9pt}
		\caption{Test benchmark functions.}
		\label{tab:table4}
		\resizebox{\textwidth}{!}{\begin{tabular}{|l|c|c|c|}
				\hline
				Test functions&Feasible spaces &d&$f_{min}$\\
				\hline
				$f_1 (x) = \sum_{i=1}^{d} x_i^2$&$[-100, 100]$&30&0\\
				\hline
				$f_2 (x) = \max_i\{|x_i|, 1\le i \le d\} $&$[-100, 100]$&30&0\\
				\hline
				$f_3 (x) = \sum_{i=1}^{d} (\sum_{j=1}^{i}  x_j)^2$&$[-100, 100]$&30&0\\
				\hline
				$f_4 (x) = \sum_{i=1}^{d} [x^2_i - 10\cos (2\pi x_i) + 10]$&$[-5.12, 5.12]$&30&0\\
				\hline
				$f_5 (x) = \sum_{i=1}^{d} ix_i^4 + random[0, 1]$&$[-1.28, 1.28]$&30&0\\
				\hline				
				$f_6 (x) = \sum_{i=1}^{d} (-x_i \sin (\sqrt{|x_i|}))$&$[-500, 500]$&30&$-418.883d$\\
				\hline	
				$f_7 (x) = \frac{1}{4000} \sum_{i=1}^{d} x^2_i - \Pi_{i=1}^{d} \cos (\frac{x_i}{\sqrt{i}}) + 1$&$[-600, 600]$&30&$0$\\
			    \hline
				$f_8 (x) = \sum_{i=1}^{d-1} [100( x_i^2 - x_{i+1})^2 + (x_i -1)^2]$&$[-5, 10]$&30&0\\
				\hline
				$f_9 (x) =\sum_{i=1}^{d} |x_i| + \Pi_{i=1}^{d}  |x_i|$&$[-10, 10]$&30&0\\
				\hline
				$f_{10} (x) =\sum_{i=1}^{d} \sin (x_i) . (\sin(\frac{i.x_i^2}{\pi}))^{2m}, m = 10 $&$[0, \pi] $&30&-4.687\\
				\hline
				$f_{11} (x) = -20 \exp (-0.2 \sqrt{\frac{1}{d}\sum_{i=1}^{d}x_i^2}) -20 \exp (\frac{1}{n}\sum_{i=1}^{d} \cos (2\pi x_i)) +20 + \exp (1) $&$[-32, 32]$&30&0\\
				\hline
				$f_{12} (x) = \{ \sum_{i=1}^{d} \sin^2 (x_i) -  \exp (-  \sum_{i=1}^{d} x_i^2 )\} . \exp (- \sum_{i=1}^{d} \sin^2 \sqrt{|x_i|} ) $&$[-10, 10]$&$30$&$-1$\\[.25cm] 
		    	\hline
		\end{tabular}}
	\end{adjustwidth}
\end{table}

\begin{figure*}[!ht]
	\setlength\abovecaptionskip{0pt}
	\setlength\belowcaptionskip{-15pt}
	\begin{adjustwidth}{-.5in}{-.5in}
		\centering
		\includegraphics{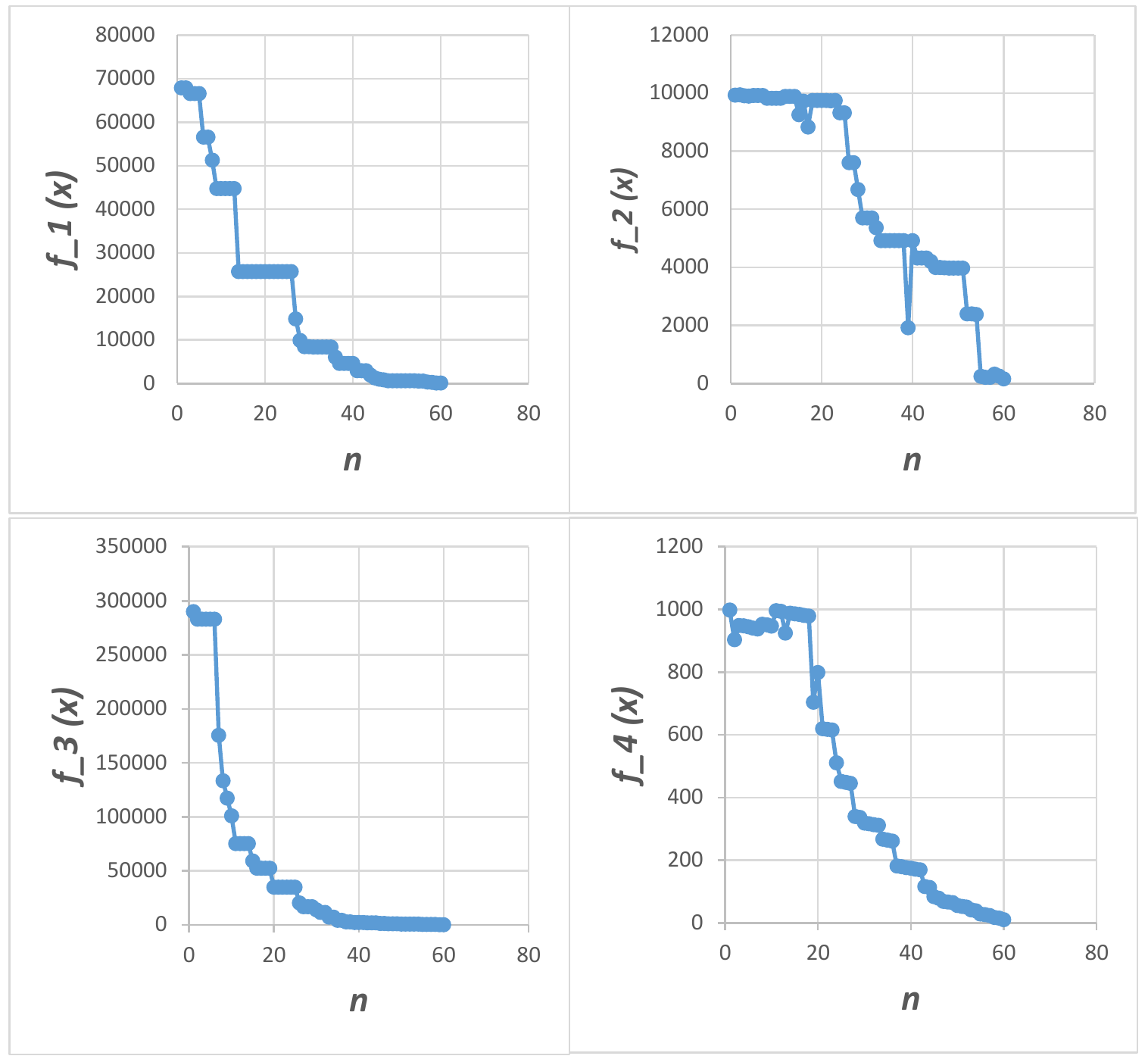}
		\caption{The results of the convergence process of $f_1(x)$, $f_2(x)$, $f_3(x)$, and $f_4(x)$ with $n =60$ and $m=10$.}
		\label{fig:figure1}
	\end{adjustwidth}
\end{figure*}

\begin{figure*}[!ht]
	\setlength\abovecaptionskip{0pt}
	\setlength\belowcaptionskip{-15pt}
	\begin{adjustwidth}{-.5in}{-.5in}
		\centering
		\includegraphics{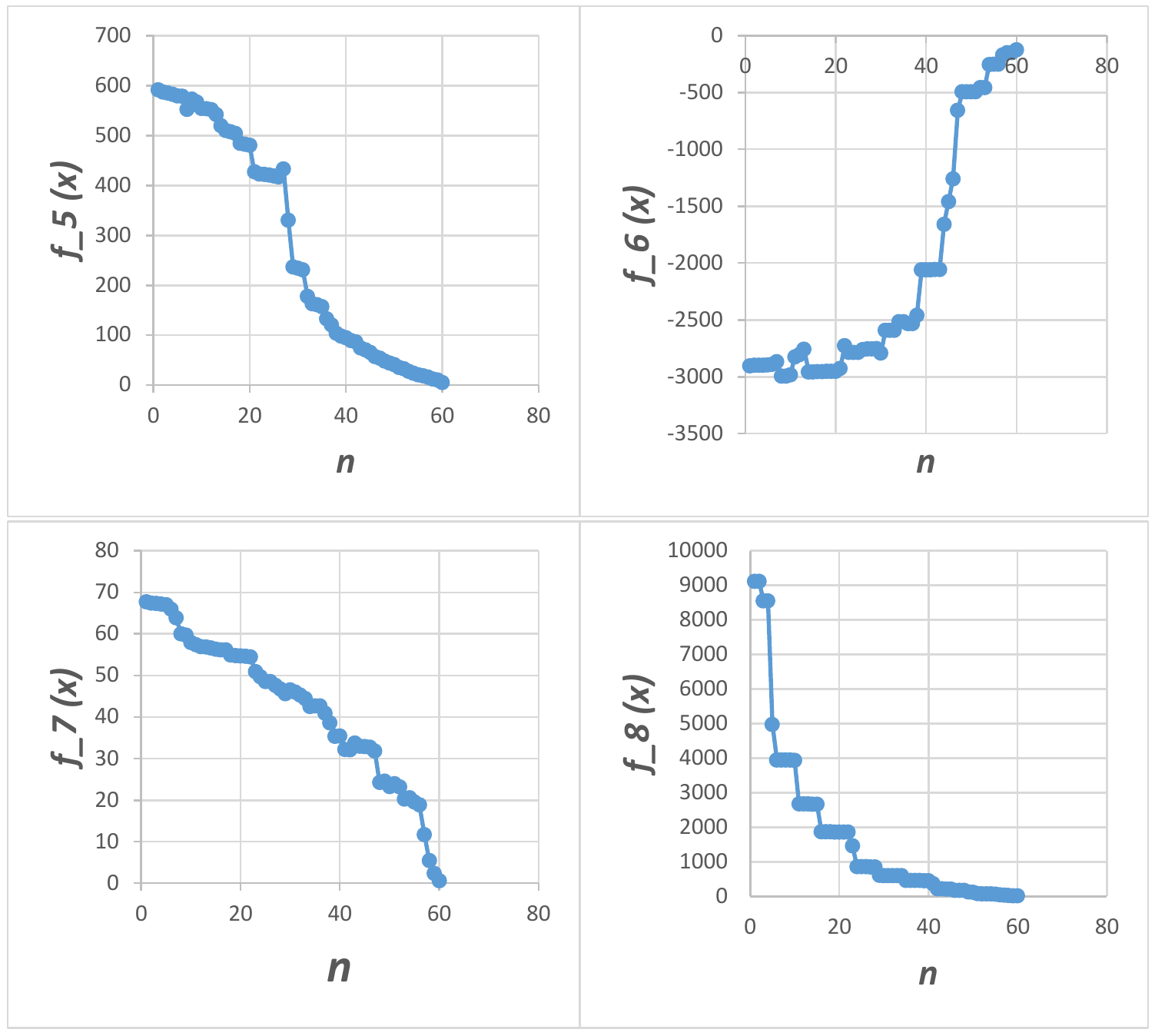}
		\caption{The results of the convergence process of $f_5(x)$, $f_6(x)$, $f_7(x)$, and $f_8(x)$ with $n =60$ and $m=20$.}
		\label{fig:figure2}
	\end{adjustwidth}
\end{figure*}

\begin{figure*}[!ht]
	\setlength\abovecaptionskip{0pt}
	\setlength\belowcaptionskip{-15pt}
	\begin{adjustwidth}{-.5in}{-.5in}
		\centering
		\includegraphics{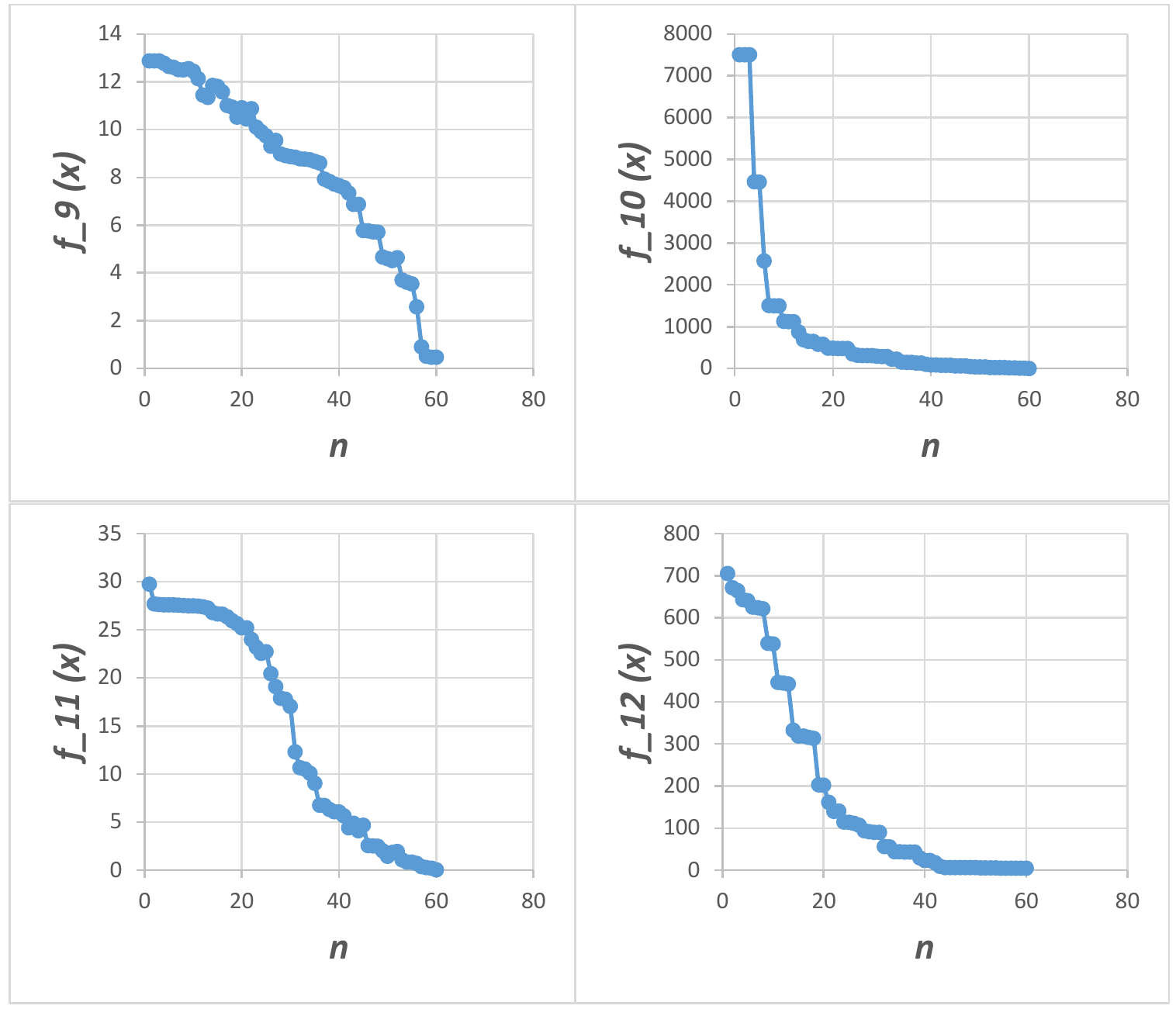}
		\caption{The results of the convergence process of $f_9(x)$ to $f_{12}(x)$ with $n =60$ and $m=10$.}
		\label{fig:figure7}
	\end{adjustwidth}
\end{figure*}

PMSAM was tested by 20 runs on each benchmark function. Over the running times, the algorithm parameters were selected to validate the convergence behavior as listed in Table ~\ref{tab:table5}. The somersault interval and eyesight are changed across the exculpation times between larger and smaller values to validate the search space increasing and decreasing. The test functions used several sets of algorithm parameters and gave the results in Figures ~\ref{fig:figure1}, ~\ref{fig:figure2}and ~\ref{fig:figure7}. From those figures, we found that the algorithm can find the feasible  optimal solution for the tested functions; moreover, maximizing the number of climb iteration based on the parallelism in PMSAM is a good factor.

\begin{table}[!ht]
	\caption{PMSAM Initialization Parameters.}
	\label{tab:table5}
	\centering
	\begin{tabular}{|l|c|}
		\hline
		\textbf{ Parameter   } &  \textbf{Value}\\
		\hline
		$n$&50 $-$ 10000\\
		\hline
		$m$&10 $-$ 100\\
		\hline
		$l$&0.0001 $-$ 0.000001\\
		\hline
		$b$&1\\
		\hline
		$[f- g]$&[-1 $-$ -10, 1 $-$ 30]\\
		\hline
		$d$&30\\
		\hline
		$P_c$&50 $-$ 500\\
		\hline
		$N_{max}$&20 $-$ 5000\\
		\hline
	\end{tabular}
\end{table}


\begin{figure}[!t]
	\centering
	\includegraphics[scale = 0.7]{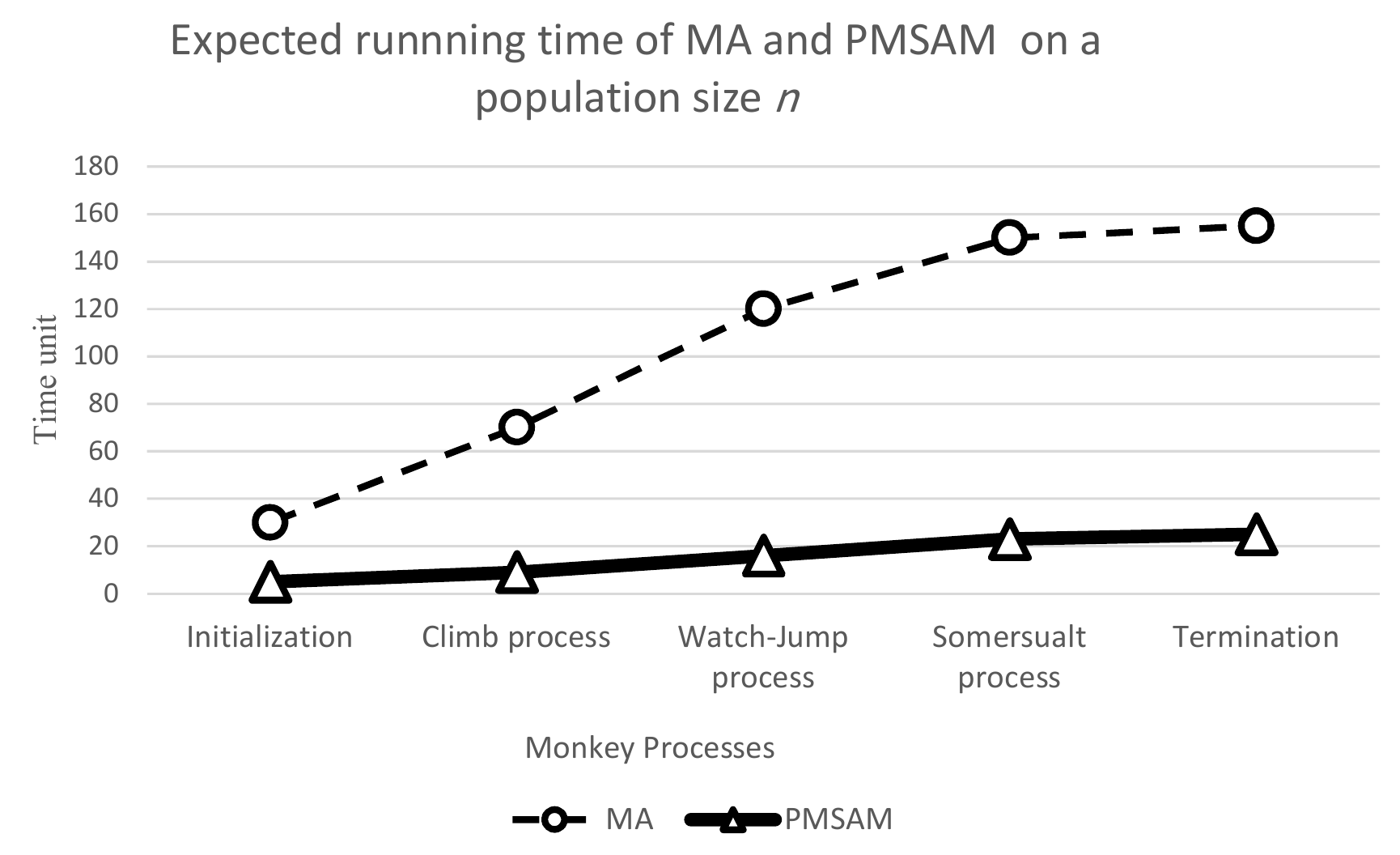}
	\caption{Expected time of PMSAM and MA running on population size $n$. The unit times are calculated based on the complexity of algorithms, where PMSAM takes a time unit to fire rule(s) over a population size $n$; as soon as, MA takes a time unit to run a step every mathematical equation over a population size $n$.}
	\label{fig:figure3}
\end{figure}

Membrane Monkey Algorithm supports convergence reliability because all monkeys can discover all search spaces by membrane migration, to update their position and choose the best optimal solutions. It runs the maximum number of search spaces and gets the exact number of optimal solutions according to the number of membranes, through algorithm processes in a time unit. MA runs on one search space and gets one optimal solution through algorithm processes in a time unit. Figure~~\ref{fig:figure3} illustrated the running time of the proposed algorithm based on Theorem ~\ref{theory7} and MA based on its time complexity measurements. Therefore, PMSAM is faster than MA according to their time complexity. It guarantees to visit search spaces, avoids local minima, and gets better probabilities for reaching the optimal solution.

The previous studies in MA used sequential processes in its development, which affected the performance of the algorithm. In this study, the performance of the base algorithm has been enhanced using the true parallelism of P systems because:
\begin{enumerate}
\item Every membrane represents a search space, and all search spaces fire climb process rules in parallel to find local optimal solutions at the same time;
\item Watch-Jump process continues to work in parallel between merged membranes, until monkeys visit all search spaces;
\item Membrane migration ensures that all monkeys will visit all search spaces to find the optimal solution;
\item Updating monkey positions is executed in parallel;
\item In nature, monkeys are searching in parallel under deterministic conditions, which is exactly simulated by the proposed algorithm;
\item Based on results of the experiment stated in Table ~\ref{tab:table6}, the algorithm provides competitive results that are compared with MA \cite{1zhao2008monkey}, and Grey Wolf Optimizer (GWO) \cite{mirjalili2014grey}. GWO is selected because it is compared to several algorithms such as particle swarm optimization, genetic algorithm, SI-based technique and a physics-based algorithm (GSA), and in addition to achieving competitive results against those algorithms \cite{mirjalili2014grey}
\end{enumerate}

\begin{table}[!ht]
	\caption{Results of tested benchmark functions after 20 runs.}
	\label{tab:table6}
	\centering
        \begin{tabular}{|l|c|c|c|c|c|c|c|c|c|c|c}
		\hline
		\multirow{2}{*}{\textbf{Func}} & \multicolumn{5}{c}{\textbf{PMSAM}} \vline& \multicolumn{2}{c}{\textbf{MA}} \vline &\multicolumn{2}{c}{\textbf{GWO}} \vline\\
		            \cline{2-10}
		 & \textbf{$m$}& \textbf{$n$}&\textbf{$P_c$}& \textbf{Mean}& \textbf{Variance}& \textbf{Mean}& \textbf{Variance}& \textbf{Mean}& \textbf{Variance}\\
		\hline
		$f_1 (x)$&	10&60&50	&1.65013E-2	&3.3460E-9&3.617E-2&3.3414E-8&4.541E-24&3.974E-11\\
		\hline
		$f_2 (x)$&20 &100&50	&2.741E-4	&1.231E-5&4.921E-4&3.342E-7&6.412E-6&1.729\\
		\hline
		$f_3 (x)$&	5&100&50	&4.027E-3&	1.0049E-7&1.371E-4&4.2231E-7&3.602E-6&6201.0217\\
		\hline
		$f_4 (x)$&	20&60&30	&1.5407E-2&	2.4010E-5&4.568E-2&3.766E-7&0.2105&2.1864\\
		\hline
		$f_5 (x)$&	10&60&30	&0.02601&	2.2471E-2&0.5381&0.015471&0.07212&1.201E-2\\
		\hline
		$f_6 (x)$&	5&100&100	&-396.045&7184.83&-403.14&9517.729&-415.39&6381.504\\
		\hline
		$f_7 (x)$&	5&60&50	&1.6010E-2&1.0283E-7&3.022E-3&1.741E-10&1.6354E-2&4.456E-4\\
		\hline
		$f_8 (x)$&	20&60&50	&0.0127&	0.006796&0.0703&0.001341&0.0478&0.001537\\
		\hline
		$f_9 (x)$&	20&60&50	&0.2504&	3.1094E-6&0.0532&1.0588E-6&5.746E-15&6.870E-4\\
		\hline
		$f_{10} (x)$&	10&100&100	&1.0071&	2.02048&1.0933&2.61087&4.05472&15.9032\\
		\hline
		$f_{11} (x)$&	10&100&30	&1.3701E-3&	2.3410&1.706E-2 &1.306E-8& 11.5304E-8 &0.543E2\\
		\hline
		$f_{12} (x)$&	20&100&100	&0.00474&3.0202&0.0721&3.01491&-0.5901&1.90927\\
		\hline
	\end{tabular}
\end{table}

In this section, we addressed the importance of P systems in the proposed algorithm from the computational power and efficiency perspectives. The previous numerical experiment introduced a proof for the efficiency of PMSAM. It was clear that the proposed algorithm depends on three variables; the number of membranes through the algorithm processes, monkey positions, and the timestamp. 
\begin{theorem}
	\label{theory7}
	The time complexity for PMSAM is $\frac{n}{m}$ with respect to $N_{max}$ and $p^{v_{g}}_{g} $, where $n$ is the population size, $m$ is number of membranes, $N_{max}$ is maximum number of iterations and $p^{v_{g}}_{g} $ is the optimal solution.
\end{theorem}
\begin{proof}
Local membranes can work at the same time based on the parallelism property, which means whatever the search spaces count is, it will not affect the timestamp. Timestamp will not increase if search spaces are increased (lines 5–8 in Algorithm ~\ref{algorithm1}).
	
Furthermore, PMSAM efficiency is determined by the timestamp. It is the most important variable because it represents time complexity state of the algorithm. Every timestamp, the rule firing happens many times in different regions over many objects. This means that algorithm processes are executed in parallel. From previous experiments, timestamp does not relate to the number of membranes (search spaces). Whatever the increase of the number of monkeys is, algorithm rules will fire in all objects at the same timestamp with the maximum number of monkeys (lines 9–20 in Algorithm ~\ref{algorithm1}). Based on numerical experiments, monkey positions faced a change without sequencing between monkeys every timestamp. That led to this result; timestamp will not be affected by the number of membranes and number of monkeys, with respect to the maximum number of monkeys (lines 5–27 in Algorithm ~\ref{algorithm1}).  Figure~~\ref{fig:figure3} shows the expected running time of PMSAM and MA over their processes. Every timestamp, MA runs on one monkey in one search space, on the contrary, PMSAM runs on $n$ monkeys and $m$ search spaces. In Figure ~~\ref{fig:figure3}, the two algorithms were traced based on the timestamp. At a time unit, one execution process is performed in the sequential mode, against $m$ execution processes in PMSAM. This means that if $m=4$, as in the previous experiment, in the climb process, one execution is done by MA compared to 4 executions by the proposed algorithm. Therefore the theorem holds.
\end{proof}

Monkey position changes across algorithm processes to find the best solution for monkey positions and the objective value. Monkeys need to discover new search spaces to find the best positions. The proposed algorithm provides a new rule in Watch-Jump process. Membrane migration does not change the monkey behavior in the real world, instead, it introduces an optimized simulation for the first step in the Watch-Jump process. It gives monkeys an advantage to avoid local minima, allowing them to discover new search spaces via deterministic rules. Furthermore, monkeys can find the best optimal solution, because they discover all search spaces in the environment before moving to somersault process.
\section{Conclusions}
\label{Conclusion and Future work}
This paper has presented an innovative method inspired by membrane computing to solve the time consuming, and sequential processing problems in MA. Its traditional method depends on a sequential mathematical model to describe monkeys movement over mountains. In this study, a novel algorithm based on P system with active membranes was innovated to model MA processes in a distributed parallel algorithm. PMSAM is working according to the P system formulation, whereas every MA process was broken down into a number of rules to perform the process over a number of objects. It simulated the real behavior of monkeys and solved time-consuming problem based on the parallelism property of the membrane computing. The results and algorithm evaluation showed how the proposed algorithm can choose a better solution. PMSAM introduces the timestamp as a new and effective stopping criterion, to control the process of the algorithm according to the allowable time.

The contribution of this study is crystallized in three points (Time complexity, nature simulation, and a better optimal solution). Time consumption problem has been solved depending on the true parallelism in P systems. Not just in Monkey Algorithm, but also in all other nature-inspired computation algorithms. The natural behavior of monkeys is simulated to open the door for formulating new nature-inspired algorithms. The proposed design provides a way to find the best optimal solution, whereas migration process keeps on searching about the optimal solution, and avoiding local minimum. It is a start point to simulate the natural parallel behavior of swarms in their mathematical models.

\end{document}